\documentclass{article}

\usepackage[preprint]{neurips_2026}
\usepackage[utf8]{inputenc}
\usepackage[T1]{fontenc}
\usepackage{hyperref}
\usepackage{url}
\usepackage{booktabs}
\usepackage{longtable}
\usepackage{array}
\usepackage{multirow}
\usepackage{amsfonts}
\usepackage{amsmath}
\usepackage{amssymb}
\usepackage{graphicx}
\usepackage{nicefrac}
\usepackage{microtype}
\usepackage{xcolor}
\usepackage{float}
\usepackage{placeins}
\usepackage{xspace}
\usepackage{tikz}

\graphicspath{{picture/}{figures/}}

\definecolor{citeblue}{RGB}{45, 105, 160}
\definecolor{codeaccent}{RGB}{68, 83, 105}
\definecolor{codefill}{RGB}{232, 238, 247}
\definecolor{websiteaccent}{RGB}{0, 127, 115}
\definecolor{websitefill}{RGB}{215, 248, 241}
\definecolor{projectaccent}{RGB}{115, 79, 210}
\definecolor{projectfill}{RGB}{238, 231, 255}
\definecolor{videoaccent}{RGB}{216, 73, 89}
\definecolor{videofill}{RGB}{255, 229, 232}
\setcitestyle{numbers,square,comma}
\hypersetup{
  colorlinks=true,
  citecolor=citeblue,
  linkcolor=citeblue,
  urlcolor=citeblue
}

\newcommand{\methodname}{MemSlides\xspace}
\newcommand{\resourcebutton}[8]{%
  \href{#5}{%
    \tikz[baseline=-0.50ex, x=1in, y=1in]{%
      \node[
        draw=#1,
        fill=#2,
        line width=0.70pt,
        rounded corners=4.6pt,
        minimum width=#6,
        minimum height=0.268in,
        inner sep=0pt
      ] (button) {};
      \node[inner sep=0pt] at ([xshift=13.1pt]button.west)
        {\includegraphics[height=#7]{#3}};
      \node[
        anchor=west,
        text=#4,
        font=\rmfamily\bfseries\fontsize{7.8pt}{8.6pt}\selectfont
      ] at ([xshift=23.0pt]button.west) {#8};
    }%
  }%
}
\newcommand{\resourcelinks}{%
  \makebox[\textwidth][c]{%
    \resourcebutton{codeaccent}{codefill}{resource-code-github-octicon.pdf}{codeaccent}{https://github.com/huohua325/Memslides}{0.69in}{0.185in}{Code}%
    \hspace{0.60em}%
    \resourcebutton{websiteaccent}{websitefill}{resource-website-globe.pdf}{websiteaccent}{https://memslides.com/}{0.85in}{0.205in}{Website}%
    \hspace{0.60em}%
    \resourcebutton{projectaccent}{projectfill}{resource-project-page.pdf}{projectaccent}{https://memslides.github.io/}{1.03in}{0.212in}{Project Page}%
    \hspace{0.60em}%
    \resourcebutton{videoaccent}{videofill}{resource-video-play.pdf}{videoaccent}{https://github.com/user-attachments/assets/a92ab49e-bc5c-4e90-8c0a-0f23b08a8857}{0.75in}{0.205in}{Video}%
  }%
}

\title{\methodname: A Hierarchical Memory Driven Agent Framework for Personalized Slide Generation with Multi-turn Local Revision}

\author{%
  Ye Jin\\
  Beijing University of Posts\\
  and Telecommunications\\
  \texttt{13681596382@bupt.edu.cn}\\
  \And
  Yangyang Xu\\
  Tsinghua University\\
  \texttt{yangyangxu@mail.tsinghua.edu.cn}\\
  \vspace{-0.85em}
  \AND
  Jun Zhu\\
  Tsinghua University\\
  \texttt{dcszj@tsinghua.edu.cn}\\
  \And
  Yibo Yang\thanks{Corresponding author.}\\
  Shanghai Jiao Tong University\\
  \texttt{yibo.yang93@gmail.com}\\
}

\begin{document}

\maketitle
\vspace{-1.55em}
\resourcelinks
\vspace{0.35em}

\begin{abstract}
Personalized presentation generation requires more than conditioning on a
current prompt or template: agents must preserve stable user preferences across
tasks, retain newly introduced preferences and constraints during multi-turn
revision, and carry out local edits reliably. We propose MemSlides, a
hierarchical memory framework for personalized presentation agents that
separates long-term memory from working memory and further divides long-term
memory into user profile memory and tool memory. User profile memory stores
intent-conditioned profiles for round-0 personalization, working memory carries
active preferences and session constraints across revision rounds, and tool
memory stores reusable execution experience for reliable localized editing.
MemSlides pairs this memory design with scoped slide-local revision, so
targeted updates act on the smallest affected region instead of repeatedly
regenerating the full deck. In controlled experiments, user profile memory
improves persona-alignment judgments on a multi-persona, multi-intent profile
bank, tool-memory injection improves closed-loop modify behavior in diagnostic
matched-pair settings, and qualitative cases illustrate working memory's ability to carryover preferences. Taken together, these results suggest that effective
personalization in presentation authoring depends on separating persistent user
profiles, session-level working memory, and reusable execution experience
across generation and localized revision.
\end{abstract}

\section{Introduction}

Automatic presentation generation aims to turn 
user requests into structured slide decks, and 
has gained increasing attention because slides are widely used, yet creating high-quality presentations remains time-consuming and cognitively demanding
\citep{sun2021d2s,maheshwari2024presentations,mondal2024personaaware,zeng2025slidetailor}.
Recent agentic systems further advance this task by 
producing 
complete decks through multi-modal or tool-based workflows
\citep{ge2025autopresent,zheng2025pptagent,xu2025pregenie,xie2025slidebot,liang2025slidegen,zheng2026deeppresenter,ozden2026arcdeck}.
Although these systems can now produce complete and visually polished decks,
they still lack persistent personalization, which is essential for generating ready-to-use slide decks because users vary in domain, purpose, style, and presentation habits.
For example, users may prefer different layouts, templates, and styles when creating slides for different purposes, such as academic presentations versus business presentations.
An effective personalized slide generation framework should build and maintain user profiles that capture users' long-term preferences in organizing, styling, and revising presentations across different intents, rather than requiring users to repeatedly specify their preferences in every interaction.



Prior work has progressively expanded the capability of presentation-generation
systems \citep{sun2021d2s,mondal2024personaaware,
bandyopadhyay2024enhancing,maheshwari2024presentations,ge2025autopresent,
xu2025pregenie,xie2025slidebot,liang2025slidegen,ozden2026arcdeck}. 
PPTAgent moves beyond text-to-slides by coupling generation with
presentation-specific evaluation \citep{zheng2025pptagent}, and DeepPresenter
introduces environment-grounded reflection for agentic presentation generation
\citep{zheng2026deeppresenter}. These systems improve general generation and
agentic refinement, but do not explicitly model user-specific personalization.
SlideTailor addresses personalization by conditioning scientific
slide generation on reference slides and task-time templates
\citep{zeng2025slidetailor}, but its personalization remains tied to provided
examples or template conditions rather than to an accumulated user profile.
These observations therefore expose a central gap: slide generation agents still lack a user-facing multi-turn dialogue process that converts revision requests into reusable preference memory and preserves local revision constraints across later turns.

\begin{figure}[!t]
  \centering
  \includegraphics[width=\linewidth]{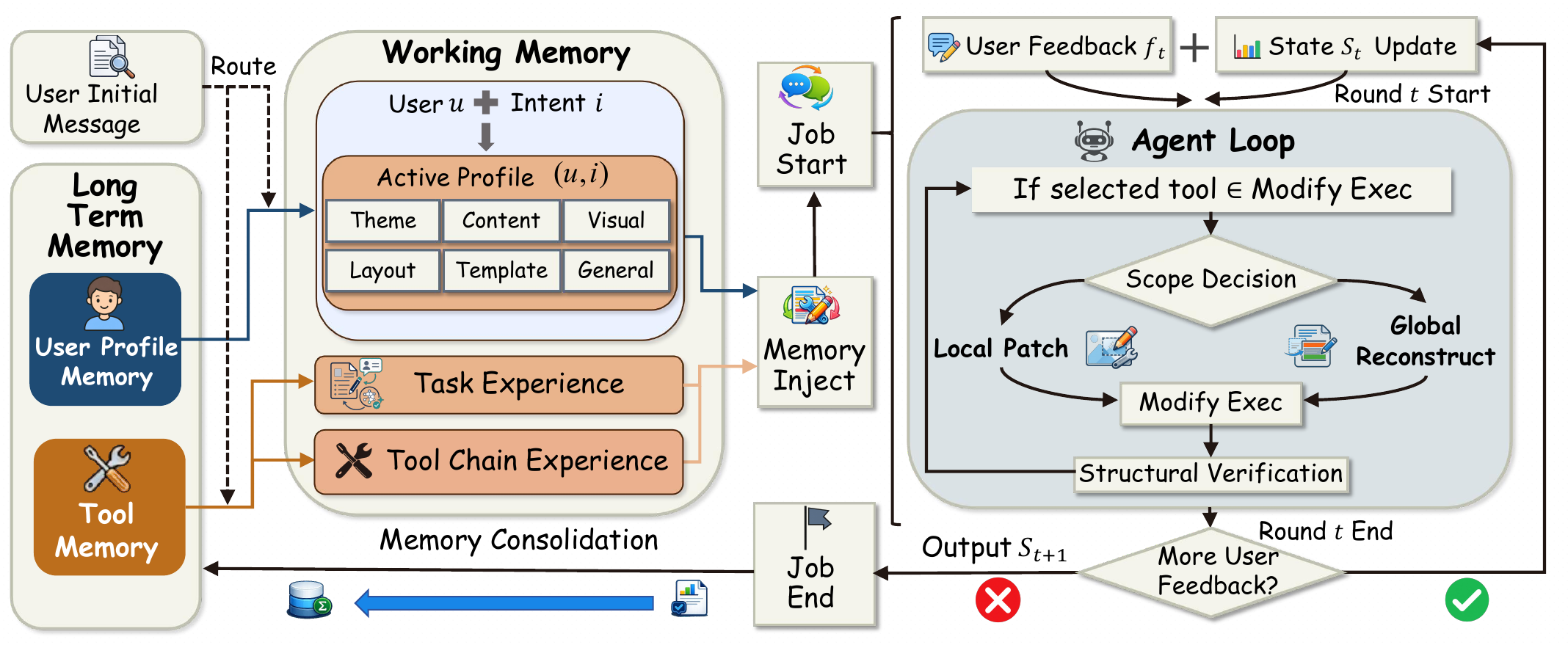}
  \vspace{-3mm}
  \caption{Overview of \methodname. The framework comprises long-term memory
  and working memory. Long-term memory stores user profile memory and tool
  memory for persistent personalization and reusable execution experience,
  while working memory carries the current session state for personalized
  generation and localized revision. In round $t$, $s_t$ is the current
  session state, $f_t$ is the user feedback, and $s_{t+1}$ is the
  updated state after \emph{Modify Exec}, the memory-guided revision executor.}
  \label{fig:pptagent_memory_workflow}
\end{figure}

This gap has two roots. First, personalization in presentation authoring is
often revealed through revision rather than fully specified before generation,
yet existing slide generation agents still handle edits by re-contextualizing or
re-generating large parts of the deck. As a result, small changes must compete
with deck state and feedback history for limited context, making multi-turn
local modification fragile. Second, current systems mainly improve
presentation-agent workflows, tools, and evaluators, but still treat
personalization as an implicit byproduct of prompting rather than a direct service enabled by
memory design. In a similar spirit to agent memory work
\citep{zhong2023memorybank,packer2023memgpt,wu2025memoryage}, personalization of 
slide generation can be largely enhanced by incorporating memory as an explicit framework rather than
an undifferentiated dialogue buffer.

To address these problems, in this paper, we present MemSlides that introduces scoped slide
locality as its revision strategy for multi-turn slide editing. Rather
than re-reading or re-writing the whole deck for each feedback turn,
\methodname projects the request onto the smallest affected slide region and
operates on a bounded repair surface. It reads only a structured snapshot of
that local surface, including its local layout structure, available selectors,
and exposed style rules, and writes back only patches scoped to explicit
selectors or to those style rules. In this way, both reading and writing stay
local by construction, reducing context pressure and unintended drift while
preserving already aligned content across turns.

Building on this modification strategy, \methodname is further integrated with a hierarchy memory
framework for personalized presentation generation. The framework has two
levels: long-term memory and working memory. Working memory maintains the
current session state and temporary feedback across revision rounds, so later
local edits can preserve active temporary memory from the same deck.
Long-term memory persists across jobs and is further divided into user profile
memory and tool memory. User profile memory is user-specific and intent-aware:
for each user and each intent, preferences are organized by various
dimensions including theme, content, visual, layout, template, and general, and are routed into working memory when each job starts. 
Tool memory captures reusable execution experience for later edits. This
hierarchy memory framework enables generating and revising presentations according to both
persistent user preferences and the active intent of the current session. The overview of \methodname is shown in Figure~\ref{fig:pptagent_memory_workflow}.

In experiments, we evaluate \methodname 
across multiple dimensions including slide deck quality, instruction satisfaction, and preference alignment. 
We develop \emph{persona-alignment judgments} as metrics to evaluate personalization alignment. 
Qualitative comparisons further illustrate the effectiveness of MemSlides in aligning with users' preferences.

Our contributions are threefold:
\begingroup
\setlength{\parskip}{0pt}
\setlength{\topsep}{2pt}
\setlength{\partopsep}{0pt}
\setlength{\itemsep}{1pt}
\setlength{\parsep}{0pt}
\begin{itemize}
    \item We introduce \methodname, a personalized presentation agent that
    supports multi-turn localized revision. By maintaining session state and
    applying targeted slide-level updates instead of repeated full-deck
    regeneration, \methodname provides the interaction substrate needed for
    learning user preferences from revision feedback.
    \item We further develop a hierarchical memory framework for \methodname consisting of
    a long-term memory and a working memory. 
    The long-term memory contains user profile memory and tool memory, allowing
    stable user preferences and reusable execution experience to persist across
    jobs, while working memory tracks session-specific constraints.
    \item We construct a multi-persona, multi-intent
    user profile bank for personalized presentation generation evaluation.
    Experiments show that user profile memory improves round-0 persona
    alignment, tool memory enhances localized modify reliability in diagnostic
    matched-pair comparisons, and working memory supports
    session-level preference carryover.
\end{itemize}
\endgroup

\section{Related Work}

\paragraph{Slide generation.}
Presentation generation has progressed from document compression and
structured summarization to LLM-based systems that emphasize audience
adaptation, editability, task-time preference inference, and visual refinement
\citep{sun2021d2s,mondal2024personaaware,zheng2025pptagent,zeng2025slidetailor,zheng2026deeppresenter}.
Presentation authoring also draws on controllable layout and design generation,
including code-like layout representations, in-context layout prompting,
layered or diffusion-based layout modeling, and visual preference modeling
\citep{tang2023layoutnuwa,lin2023layoutprompter,inoue2023layoutdm,zhang2023layoutdiffusion,peng2025designpref}.
These works improve slide quality, editability, task-level controllability, and
visual composition. Our focus is complementary: how user-specific preference
and execution memory should persist across slide-generation and revision
sessions rather than being supplied only as current-task inputs.

\paragraph{Memory and tool-using agents.}
Retrieval-augmented and external-memory language models show that stored
context can support generation
\citep{lewis2020rag,guu2020realm,khandelwal2019knnlm,borgeaud2021retro,izacard2022atlas,wang2023longmem}.
These studies focus on
persistent memory, reflection, structured
updates, and long-term/short-term memory management
\citep{zhong2023memorybank,packer2023memgpt,park2023generativeagents,xu2025amem,chhikara2025mem0,wang2025mirix,kang2025memoryos,wu2025memoryage,yu2026agemem}.
Tool-using and reflective agents further establish patterns for interleaving
reasoning with actions, using APIs or modular tools, coordinating execution,
and learning from feedback
\citep{yao2023react,shinn2023reflexion,schick2023toolformer,patil2023gorilla,qin2023toolllm,li2023apibank,karpas2022mrkl,wu2023autogen,wang2023voyager,madaan2023selfrefine,nakano2021webgpt,ahn2022saycan,wang2022selfconsistency,yao2023treeofthoughts,hu2026agentictooluse}.
In contrast, \methodname targets personalized presentation authoring, where
memory must distinguish user preference from execution experience and preserve
the intended local edit scope during multi-turn revision.

\paragraph{Personalized generation and evaluation.}
Personalized generation has evolved from explicit persona conditioning to
profile- and history-aware generation
\citep{li2016persona,wu2021splitmemory,salemi2023lamp,mysore2023pearl,jiang2025personamem}.
Recent surveys and alignment work characterize personalization as an agentic,
retrieval-aware, and preference-sensitive problem
\citep{zhang2024personalizationllm,li2025personalizationragagent}. In visual
domains, personalized visualization recommendation and DesignPref show that
persistent expressive or design preferences can be learned from user history
\citep{qian2022personalizedvis,peng2025designpref}. In PPT generation,
Persona-Aware-D2S and SlideTailor personalize slides through audience
specifications, examples, or templates provided for the current task, whereas
we study preferences accumulated across tasks and retained during multi-turn
revision. Our evaluation follows rubric-based and pairwise LLM-as-judge
protocols in a controlled memory/no-memory setting
\citep{liu2023geval,zheng2023llmasjudge,kim2023prometheus,zhu2023judgelm,li2024arenahard}.

\section{MemSlides}

\subsection{Problem Formulation}

We formulate personalized presentation generation as a stateful, multi-turn authoring problem rather than a one-shot source-to-slides conversion task
\citep{sun2021d2s,zheng2025pptagent,zeng2025slidetailor,zheng2026deeppresenter}.
Given source material $x$, a user profile memory $P_u$, and an optional task-time template $\tau$, the system first produces an initial deck:
\begin{equation}
S_0 = G_{\mathrm{init}}(x, P_u, \tau).
\end{equation}
We refer to this initial generation stage as \emph{round-0}. After receiving user feedback $f_t$ at revision round $t$, the system updates a session state
$z_t$ and edits the current deck accordingly:
\begin{equation}
z_t = U(z_{t-1}, f_t; S_{t-1}), \qquad
S_t = G_{\mathrm{edit}}(S_{t-1}, x, P_u, \tau, z_t), \quad t \ge 1.
\end{equation}
Here, $z_t$ stores feedback-derived constraints and edit intentions that are active within the current authoring session. The objective is not only to improve the first draft, but also to move the deck toward the user's intended presentation while preserving slides that are already well aligned.

This formulation separates three personalization signals with different lifetimes. The user profile memory $P_u$ captures recurring cross-job preferences, such as preferred themes, visual styles, layout habits, content density, and general design conventions. The task-time template $\tau$, when provided, serves as a deck-local design constraint for the current job. The session state $z_t$ captures temporary, turn-specific requirements, such as changing the color of new slide titles or compressing text after a particular section.
When these signals conflict, explicit session feedback takes precedence for the current deck, followed by the task-time template and then the user profile memory.

The temporal structure also determines the operational scope of revision. We define a \emph{job} as one complete deck-authoring session, a \emph{round} (turn) as one user-triggered revision within that job, and an \emph{operation} as a concrete edit or tool call. Revision requests may be local to a single slide, propagate across multiple slides, or modify the deck structure itself.
Therefore, repeatedly regenerating the full deck is undesirable: it can overwrite already aligned content, introduce unnecessary drift, and increase context pressure. A stateful revision system should instead select the minimal effective scope for each request while carrying forward relevant session constraints.

\begin{figure}[!t]
  \centering
  \includegraphics[width=\linewidth]{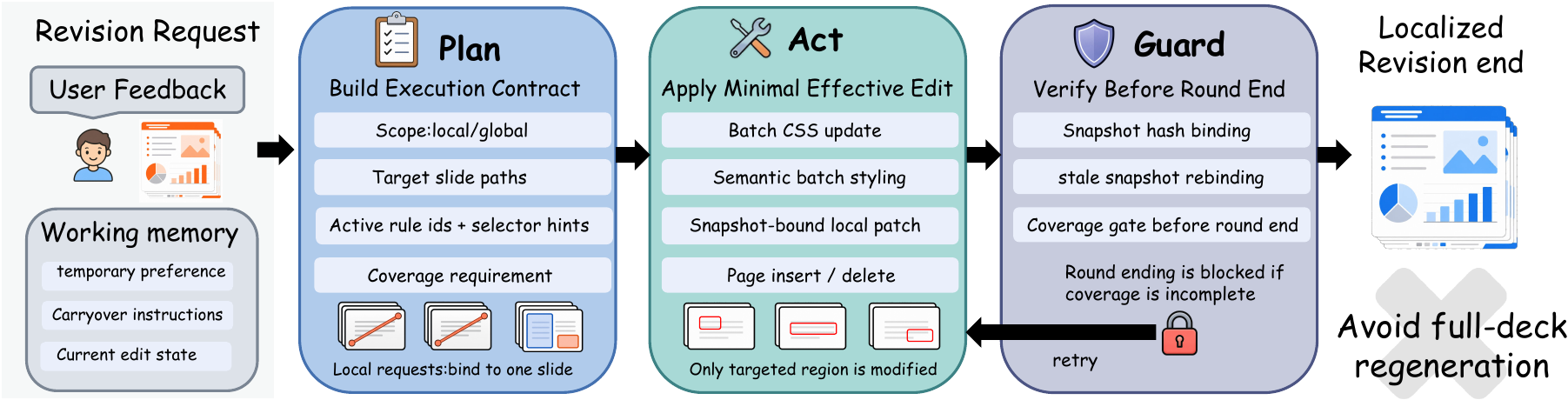}
  \vspace{-2mm}
  \caption{Localized modify execution in \methodname. Working memory supplies
  active temporary preferences, carryover instructions, current edit state, and
  buffered tool-memory signals to the Plan--Act--Guard pipeline. Plan builds an
  execution contract, Act applies the minimal effective edit, and Guard verifies
  coverage before finalizing the localized update.}
  \label{fig:localized_tool_pipeline}
\end{figure}

\subsection{Multi-Turn Localized Modify Execution}

\textbf{Plan.} \methodname converts each revision request into an explicit
execution contract rather than leaving scope control implicit in the prompt. The
contract records the inferred scope, target slide paths, active rule identifiers,
selector hints, and whether target coverage is required. Local requests bind to
the resolved slide; deck-level rules expand coverage to all slides; and hybrid
requests preserve both the global rule and the local exception. Future-only
rules created by an insertion request are not incorrectly forced onto existing
slides, which avoids turning a structural edit into an unnecessary deck-wide
rewrite.

\textbf{Act.} The executor chooses editing tools according to this contract and
the available slide structure. If target slides share an explicit selector, it
prefers a batch CSS update; if the change targets common semantics such as
titles, body text, or footers across structurally different slides, it uses
semantic batch styling; if a single existing slide needs content or local
structure changes, it reads a layout-first repair surface and applies
non-empty patch operations to snapshot targets or exposed rules. Page insertion
and deletion remain explicit page-level operations, while whole-slide rewriting
is restricted to new slides or controlled recovery after a corrupted or
uninspectable state.

\textbf{Guard.} The protocol treats completion as a checked state, not merely as
the model deciding to stop. Patch calls are bound to the snapshot content hash;
stale snapshots trigger rebinding hints rather than immediate full rewrite; and
every patch must specify concrete operations derived from repair candidates,
rules, or targets. When coverage is required, premature \texttt{finalize} calls
are blocked until all target slides are modified or explicitly confirmed
compliant. These guards make localized revision a constrained execution process:
the agent can update the intended deck region while reducing repeated
re-contextualization, uncontrolled scope expansion, and drift in already aligned
slides.

Working memory is the session-scoped state layer that makes
Plan--Act--Guard multi-turn rather than single-shot. It stores active temporary
preferences from earlier feedback, carryover instructions that remain valid for
the current deck, and edit-state records such as resolved targets, coverage
status, and snapshot rebinding hints. It also buffers round-level tool-memory
signals before transferable experiences are consolidated into long-term tool
memory. Plan reads this state to construct the execution contract, Act uses it
to restrict edits to the intended region, and Guard updates it after
verification or rebinding.

As illustrated in Figure~\ref{fig:localized_tool_pipeline}, localized
modification gives the agent an editable substrate, but personalization still requires separating signals by lifetime and use. Rather
than using one homogeneous buffer
\citep{lewis2020rag,guu2020realm,khandelwal2019knnlm,borgeaud2021retro,izacard2022atlas,wang2023longmem},
\methodname instantiates hierarchy memory through two modules: user profile
memory decides what the deck should reflect, and tool memory supports how edits
are executed, following agent-memory work on persistent and session state
\citep{zhong2023memorybank,packer2023memgpt,park2023generativeagents,xu2025amem,chhikara2025mem0,wang2025mirix,kang2025memoryos,wu2025memoryage,yu2026agemem}.

\subsection{User Profile Memory for Personalization}

\begin{figure}[!t]
  \centering
  \includegraphics[width=1.0\linewidth]{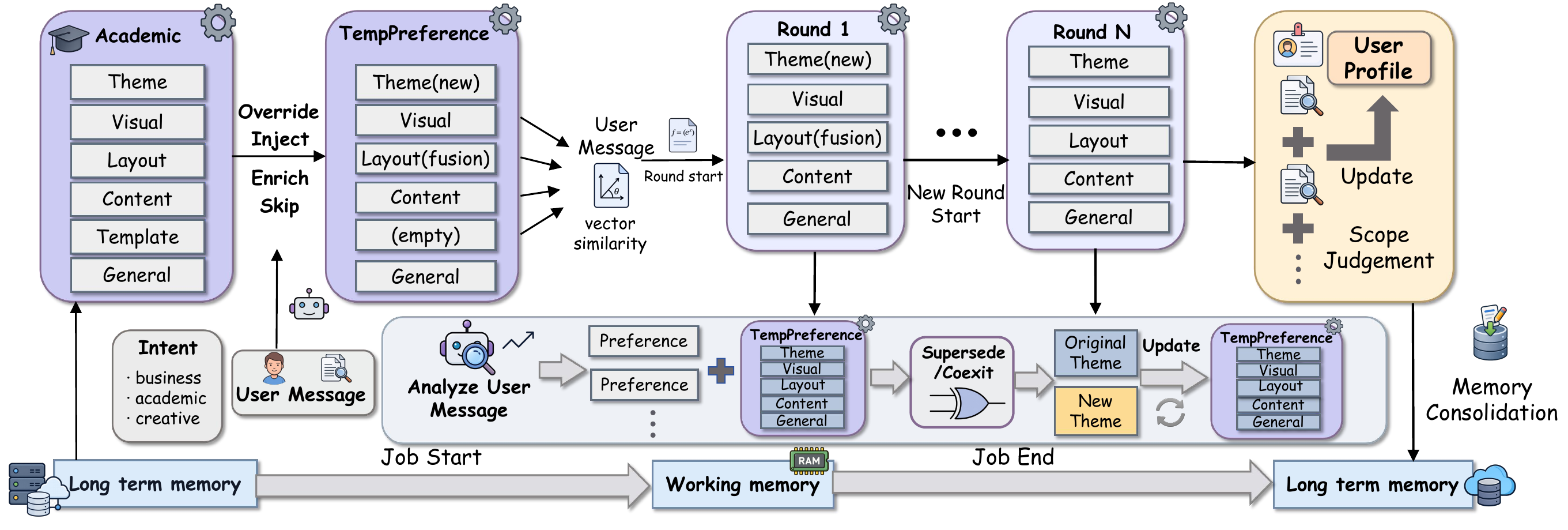}
  \vspace{-3mm}
  \caption{User profile memory lifecycle in \methodname. Long term memory stores
  intent-conditioned user profile memory accumulated across jobs. At job start,
  user profile memory items are routed into active temporary memory by comparing them with the
  current user request: compatible preferences coexist, explicit conflicts are
  superseded, and only the active subset guides generation. At job end, stable interaction signals are consolidated
  back into the user profile memory.}
  \label{fig:preference_memory_lifecycle}
\end{figure}

User profile memory governs personalization in \methodname.
Instead of using personalization as a static prompt prefix, we represent it as a
persistent user profile plus active temporary memory for the current session:
\begin{equation}
\label{eq:preference_memory_state}
\mathcal{M}^{\mathrm{pref}}_t = \big(P_u, A_t\big),
\end{equation}
where $P_u$ is user $u$'s long-term profile and $A_t$ is the active temporary
memory at revision round $t$. Stored profile items are organized by intent and
presentation dimensions, while only the reconciled subset enters $A_t$. This
separation is the lifecycle shown in
Figure~\ref{fig:preference_memory_lifecycle}: user profile memory provides
cross-job personalization priors, routing turns the intent-matched and
request-compatible items into active temporary memory for the current deck, and
job-end consolidation writes back only stable interaction signals. The figure
therefore explains why profile memory is not injected as a static prompt block:
it is selected, reconciled with the current request, used during round-0 and
later revisions, and then updated after the round.

Before round-0 generation, \methodname selects the profile bucket for the
current intent $i_0$, extracts constraints from the request $q_0$, and routes
compatible items into working memory:
\begin{equation}
\label{eq:profile_routing}
\begin{aligned}
\tilde P_u = \mathcal{S}\!\left(P_u, i_0\right),
\qquad
C_0 = \mathcal{E}\!\left(q_0\right),
\qquad
A_0 = \mathcal{R}\!\left(\tilde P_u, C_0\right).
\end{aligned}
\end{equation}
Here, $\mathcal{S}$ retrieves the intent-matched profile bucket, $\mathcal{E}$
extracts request constraints, and $\mathcal{R}$ reconciles them. Explicit
request conflicts supersede the corresponding profile item for the current deck;
compatible preferences coexist as active memory.

During revision, active temporary memory evolves with feedback:
\begin{equation}
\label{eq:working_memory_update}
A_t = \mathcal{U}\!\left(A_{t-1}, r_t\right),
\end{equation}
where $r_t$ is the revision request at round $t$. The update operator
$\mathcal{U}$ appends newly exposed preferences, supersedes true conflicts, and
keeps non-conflicting items active for later rounds.

At job end, long-term user profile memory is updated by consolidation rather
than by directly promoting every temporary item:
\begin{equation}
\label{eq:profile_consolidation}
P_u^{+} = \mathcal{C}\!\left(P_u, H\right),
\end{equation}
where $H$ denotes the user messages in the current job and $\mathcal{C}$ is
intent-aware profile consolidation.
This prevents transient requests from being stored as persistent preferences, while allowing stable, transferable signals to improve future personalization.

\subsection{Tool Memory for Reliable Editing}

Tool memory addresses the execution side of personalized editing. Even when the
target preference is correct, the agent may repeat ineffective trials or
re-trigger known tool misuse. Prior tool-using and reflective agents motivate
interleaving reasoning, tool calls, and feedback
\citep{yao2023react,shinn2023reflexion,schick2023toolformer,qin2023toolllm,madaan2023selfrefine}.
In our setting, the remembered execution unit is tied to slide-edit scope and
verification rather than to raw interaction history.

\begin{figure}[!t]
  \centering
  \includegraphics[width=\linewidth]{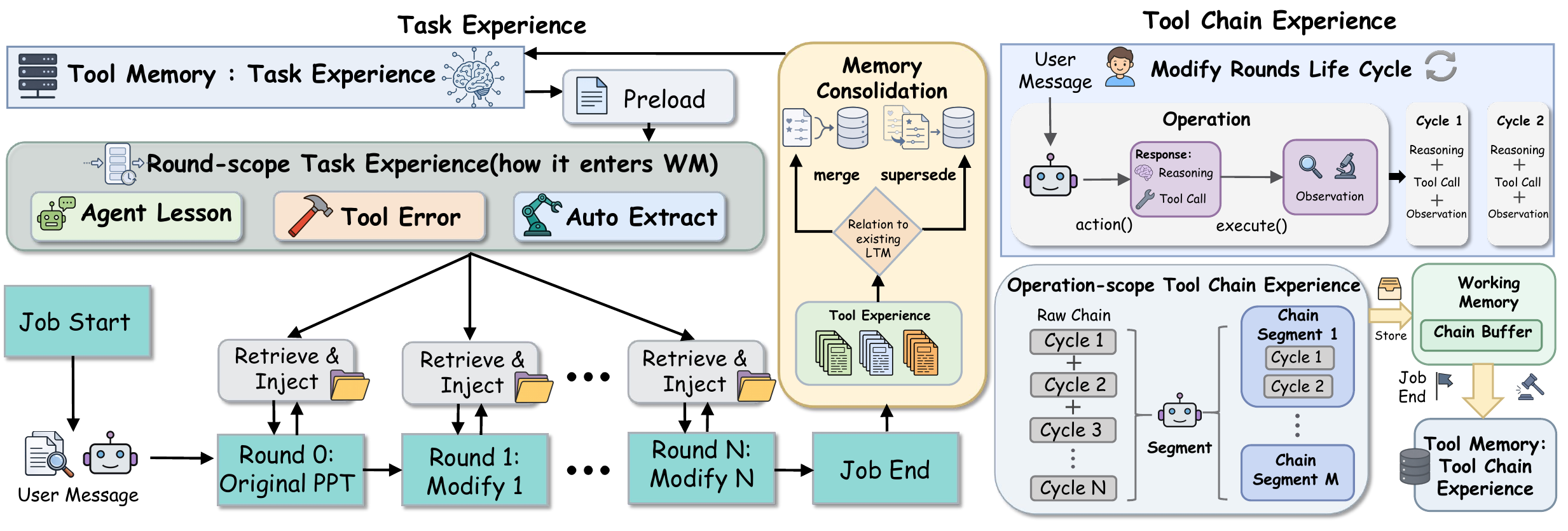}
  \vspace{-4mm}
  \caption{Tool memory flow in \methodname. Round-scope task experience is
  available at job start, enters working memory during modify rounds, and is
  updated through agent lessons, tool-error summaries, and automatically
  extracted patterns before memory consolidation. Operation-scope tool-chain
  experience records raw reasoning--tool--observation chains as compact
  fragments that are retrieved and injected before similar future tool calls.}
  \label{fig:tool_memory_lifecycle}
\end{figure}

In \methodname, tool memory is organized two execution flows that match
the temporal granularity of localized editing
(Figure~\ref{fig:tool_memory_lifecycle}). 
Round-scope task experience is available when a modify job starts and is
buffered in working memory across later modify rounds; after each round, agent
lessons, tool-error summaries, and automatically extracted patterns can update
this pool. Operation-scope tool-chain experience is finer grained: raw
reasoning--tool--observation chains are segmented into reusable chain fragments,
indexed by operation context, and retrieved before similar future tool calls.

We represent this execution support at revision round $t$ and operation $k$ as
\begin{equation}
\label{eq:tool_memory_state}
\mathcal{M}^{\mathrm{tool}}_{t,k} = \big(E_t^{\mathrm{round}}, E_{t,k}^{\mathrm{op}}\big),
\end{equation}
where $E_t^{\mathrm{round}}$ denotes round-scope execution experience and $E_{t,k}^{\mathrm{op}}$ denotes operation-scope tool chain experience for the $k$-th operation. Tool memory does not define what the deck should look like; it helps the agent execute that objective with fewer repeated errors, less backtracking, and more reliable local verification. Transferable experiences are consolidated into long-term tool memory at job end.

\section{Experiments}

\vspace{-1.2ex}
We evaluate whether \methodname improves personalized presentation generation
and multi-turn localized revision. The main text reports personalization results
on a controlled multi-persona, multi-intent user profile bank, a
DeepPresenter-style general-quality check on the same decks, and a diagnostic
matched-pair modify evaluation for localized revision. Protocol details,
baseline conditions, and profile-bank construction are provided in
Appendices~\ref{app:evaluation_protocol_details},
\ref{app:baseline_provenance}, \ref{app:profile_bank_construction},
and~\ref{app:deeppresenter_quality_eval}, with protocol, profile-bank, and
compute summaries in Appendix Tables~\ref{tab:appendix_eval_protocol_summary},
\ref{tab:appendix_profile_library_catalog}, and
\ref{tab:appendix_compute_accounting}; table captions define the reported
metrics.

\subsection{Main Results}

\subsubsection{Personalization}

\begin{table}[t]
\centering
\small
\setlength{\tabcolsep}{6pt}
\renewcommand{\arraystretch}{1.0}
\caption{Persona-alignment judgments for first-pass generation. Scores are
averaged over three personas on a 0--10 scale; higher is better. Bold marks the
best score per metric.}
\label{tab:profile_memory_v6_bestof_main}
\begin{tabular}{clcccc}
\toprule
Framework & Model & Content $\uparrow$ & Structure $\uparrow$ & Visual $\uparrow$ & Specificity $\uparrow$ \\
\midrule
\multirow{3}{*}{\centering DeepPresenter} & GPT-5 & 6.22 & 7.56 & 5.76 & 5.89 \\
& GLM-5 & 6.67 & 7.61 & 5.28 & 7.22 \\
& Gemini 3.1 Pro & 6.89 & 8.00 & 6.78 & 7.44 \\
\midrule
\multirow{3}{*}{\centering SlideTailor} & GPT-5 & 6.78 & 6.00 & 6.39 & 6.33 \\
& GLM-5 & 4.44 & 4.89 & 4.00 & 3.89 \\
& Gemini 3.1 Pro & 4.48 & 5.00 & 4.03 & 4.67 \\
\midrule
\multirow{3}{*}{\centering MemSlides (Ours)} & GPT-5 & 7.11 & 7.33 & 6.00 & 6.67 \\
& GLM-5 & \textbf{9.00} & \textbf{8.78} & \textbf{8.56} & \textbf{8.89} \\
& Gemini 3.1 Pro & 7.77 & 8.64 & 8.24 & 8.56 \\
\bottomrule
\end{tabular}
\end{table}

User profile memory improves round-0 persona alignment across multiple
dimensions. Under the persona-alignment judgments, \textsc{Ours} achieves
all-column wins over both baselines on GLM-5 and Gemini 3.1 Pro. With GPT-5, it
remains ahead of SlideTailor on Content, Structure, and Specificity, and ahead
of DeepPresenter on Content, Visual, and Specificity, while DeepPresenter has
slightly higher Structure (Table~\ref{tab:profile_memory_v6_bestof_main};
Appendix Table~\ref{tab:profile_memory_v6_bestof_gpt_appendix} gives
per-persona GPT-5 details).

\begin{table}[t]
\centering
\small
\setlength{\tabcolsep}{5.2pt}
\renewcommand{\arraystretch}{1.0}
\caption{General-quality evaluation on the
shared three-profile suite.  Constraint, Content, Style, and Avg. use a 1--5 scale;
Diversity is a suite-level normalized DINOv2-Vendi score. The rows compare
independent generated decks from DeepPresenter, SlideTailor, and \methodname
under the same evaluation protocol.}
\label{tab:profile_memory_v6_ppteval_main}
\begin{tabular*}{\textwidth}{@{\extracolsep{\fill}}clccccc@{}}
\toprule
Framework & Model & Constraint $\uparrow$ & Content $\uparrow$ & Style $\uparrow$ & Avg. $\uparrow$ & Diversity $\uparrow$ \\
\midrule
\multirow{3}{*}{\centering DeepPresenter}
& GPT-5 & 4.83 & 3.50 & 3.63 & 3.99 & 0.387 \\
& Gemini 3.1 Pro & 4.17 & 3.33 & 4.00 & 3.83 & 0.370 \\
& GLM-5 & 4.00 & 3.57 & 4.00 & 3.86 & 0.366 \\
\midrule
\multirow{3}{*}{\centering SlideTailor}
& GPT-5 & 3.83 & 2.93 & 4.03 & 3.60 & 0.399 \\
& Gemini 3.1 Pro & 3.83 & 3.20 & 4.00 & 3.68 & 0.364 \\
& GLM-5 & 3.83 & 2.97 & 4.00 & 3.60 & 0.348 \\
\midrule
\multirow{3}{*}{\centering MemSlides ({Ours})}
& GPT-5 & \textbf{5.00} & \textbf{3.60} & 3.90 & \textbf{4.17} & 0.380 \\
& Gemini 3.1 Pro & 3.33 & 3.37 & \textbf{4.10} & 3.60 & \textbf{0.463} \\
& GLM-5 & 3.83 & 3.34 & 4.03 & 3.74 & 0.391 \\
\bottomrule
\end{tabular*}
\end{table}

\begin{table}[t]
\centering
\small
\setlength{\tabcolsep}{6pt}
\renewcommand{\arraystretch}{0.9}
\newcommand{\toolmemorymultiheader}[1]{\shortstack[c]{#1}}
\providecommand{\cmark}{\textcolor{green!50!black}{\checkmark}}
\providecommand{\xmark}{\textcolor{red!70!black}{\(\times\)}}
\newcommand{\toolmemoryyes}{\cmark}
\newcommand{\toolmemoryno}{\xmark}
\caption{Tool-memory ablation on nine diagnostic modify pairs. Completion and
verification are higher-is-better; time and ratio are lower-is-better.}
\label{tab:tool_memory_table1_redesign}
\begin{tabular*}{\textwidth}{@{\extracolsep{\fill}}lccccc@{}}
\toprule
Model & \toolmemorymultiheader{Memory\\Injected} &
\toolmemorymultiheader{Closed-Loop\\Completion $\uparrow$} &
\toolmemorymultiheader{Strict\\Verify $\uparrow$} &
\toolmemorymultiheader{First Correct\\Edit (s) $\downarrow$} &
\toolmemorymultiheader{Core Tool\\Time Ratio $\downarrow$} \\
\midrule
\multirow{2}{*}{GPT-5} & \toolmemoryyes{} & \textbf{1.000} & \textbf{0.646} & \textbf{211.3} & \textbf{0.740$\times$} \\
 & \toolmemoryno{} & 0.667 & 0.294 & 234.2 & 1.000$\times$ \\
\midrule
\multirow{2}{*}{GLM-5} & \toolmemoryyes{} & \textbf{1.000} & \textbf{0.488} & \textbf{195.9} & \textbf{0.344$\times$} \\
 & \toolmemoryno{} & 0.889 & 0.434 & 500.9 & 1.000$\times$ \\
\midrule
\multirow{2}{*}{Gemini 3.1 Pro} & \toolmemoryyes{} & 0.889 & \textbf{0.469} & \textbf{309.9} & \textbf{0.137$\times$} \\
 & \toolmemoryno{} & 0.889 & 0.201 & 968.2 & 1.000$\times$ \\
\midrule
\multirow{2}{*}{Overall} & \toolmemoryyes{} & \textbf{0.963} & \textbf{0.534} & \textbf{242.5} & \textbf{0.327$\times$} \\
 & \toolmemoryno{} & 0.815 & 0.310 & 609.5 & 1.000$\times$ \\
\bottomrule
\end{tabular*}
\end{table}

On the same generated decks, the DeepPresenter-style quality metrics in
Table~\ref{tab:profile_memory_v6_ppteval_main} check whether persona gains
remain compatible with ordinary presentation quality. The results show that
\methodname improves personalization alignment while maintaining competitive
PPT generation quality.


\subsubsection{Localized Revision}

In this diagnostic matched-pair modify setting, tool-memory injection is
associated with stronger closed-loop completion and post-edit verification while
reducing time to the first correct edit and core tool time
(Table~\ref{tab:tool_memory_table1_redesign}). Since each matched pair changes
only tool-memory injection, these process gains support the combined
tool-memory pathway for reliable multi-turn modification, while the W-L-T-NA
counts expose pair-level heterogeneity; Appendix
Tables~\ref{tab:tool_memory_table1_pair_detail} and
\ref{tab:tool_memory_paired_robustness} provide pair-level details and the
paired robustness check.

\begin{figure}[!t]
  \centering
  \includegraphics[width=\linewidth]{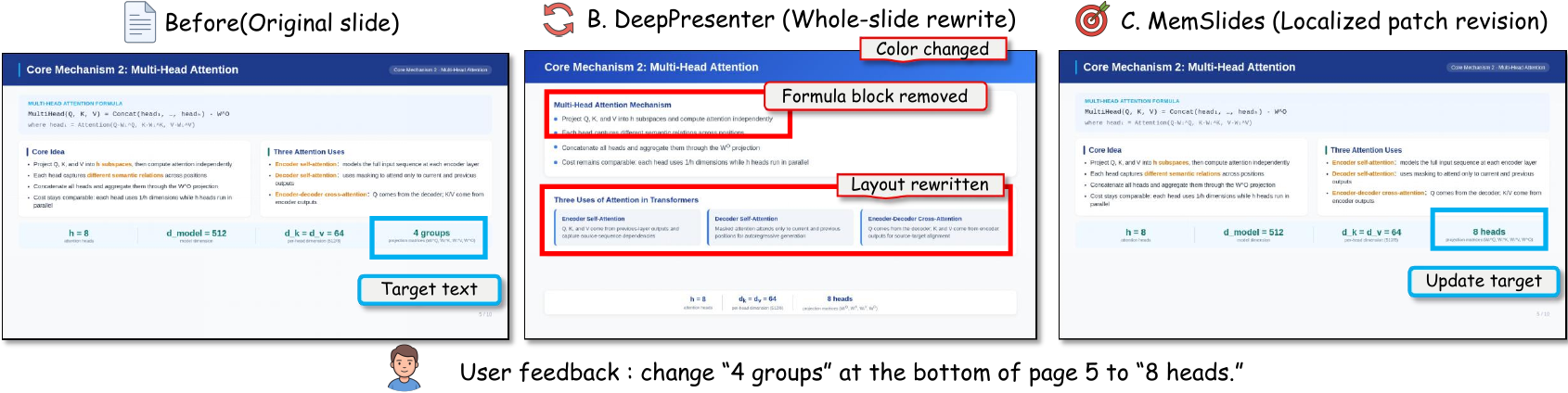}
  \vspace{-4mm}
  \caption{Qualitative illustration of localized modify execution. Given the
  same local edit request, DeepPresenter can satisfy the target change while
  altering non-target regions of the slide. \methodname instead applies a
  targeted patch to the requested element, preserving already aligned slide
  content.}
  \label{fig:localized_modify_example}
\end{figure}

The same locality-sensitive setting is illustrated qualitatively in
Figure~\ref{fig:localized_modify_example}, which contrasts a broader edit
footprint with a targeted patch on the requested element; Appendix
Figure~\ref{fig:appendix_tool_memory_localized_editing} shows a complementary
process trajectory.

\FloatBarrier

\subsection{Analysis and Discussion}

\noindent\textbf{User profile memory yields broad, not marginal, alignment gains.}
Table~\ref{tab:profile_memory_v6_bestof_main} shows a consistent but not
uniform pattern across model families. With GLM-5 and Gemini 3.1 Pro,
\methodname leads both baselines on all four persona-alignment dimensions. With
GPT-5, the gains are strongest on Content and Specificity, while the two
baselines retain isolated advantages on Structure and Visual. Averaged across
model families, \methodname improves over DeepPresenter by 1.37 points on
Content, 0.53 on Structure, 1.66 on Visual, and 1.19 on Specificity, and
improves over SlideTailor by 2.73, 2.95, 2.79, and 3.08 points on the same
dimensions.

\noindent\textbf{The gains are planning-level persona gains.}
The strongest evidence is the joint movement of Structure and Specificity.
Structure excludes template retrieval accuracy, while Specificity uses
distractor personas; moreover, judges do not see the original prompt or parsed
intent. Thus the improvement cannot be read as merely better template matching
or generic visual polish. It indicates that routed long-term profiles help
decide page roles, ordering, layout fit, evidence emphasis, and persona-distinct
framing before round-0 generation.

\noindent\textbf{Persona gains remain compatible with general deck quality.}
Table~\ref{tab:profile_memory_v6_ppteval_main} shows that \textsc{Ours} obtains
the best Avg. on GPT-5 and remains competitive on GLM-5, while Gemini 3.1 Pro
shows the best Style and Diversity but lower Constraint. These results do not
claim uniform dominance on every quality metric; they show that the
persona-alignment improvements in Table~\ref{tab:profile_memory_v6_bestof_main}
are not simply a trade-off against ordinary presentation quality.

\begin{figure}[!t]
  \centering
  \includegraphics[width=\linewidth]{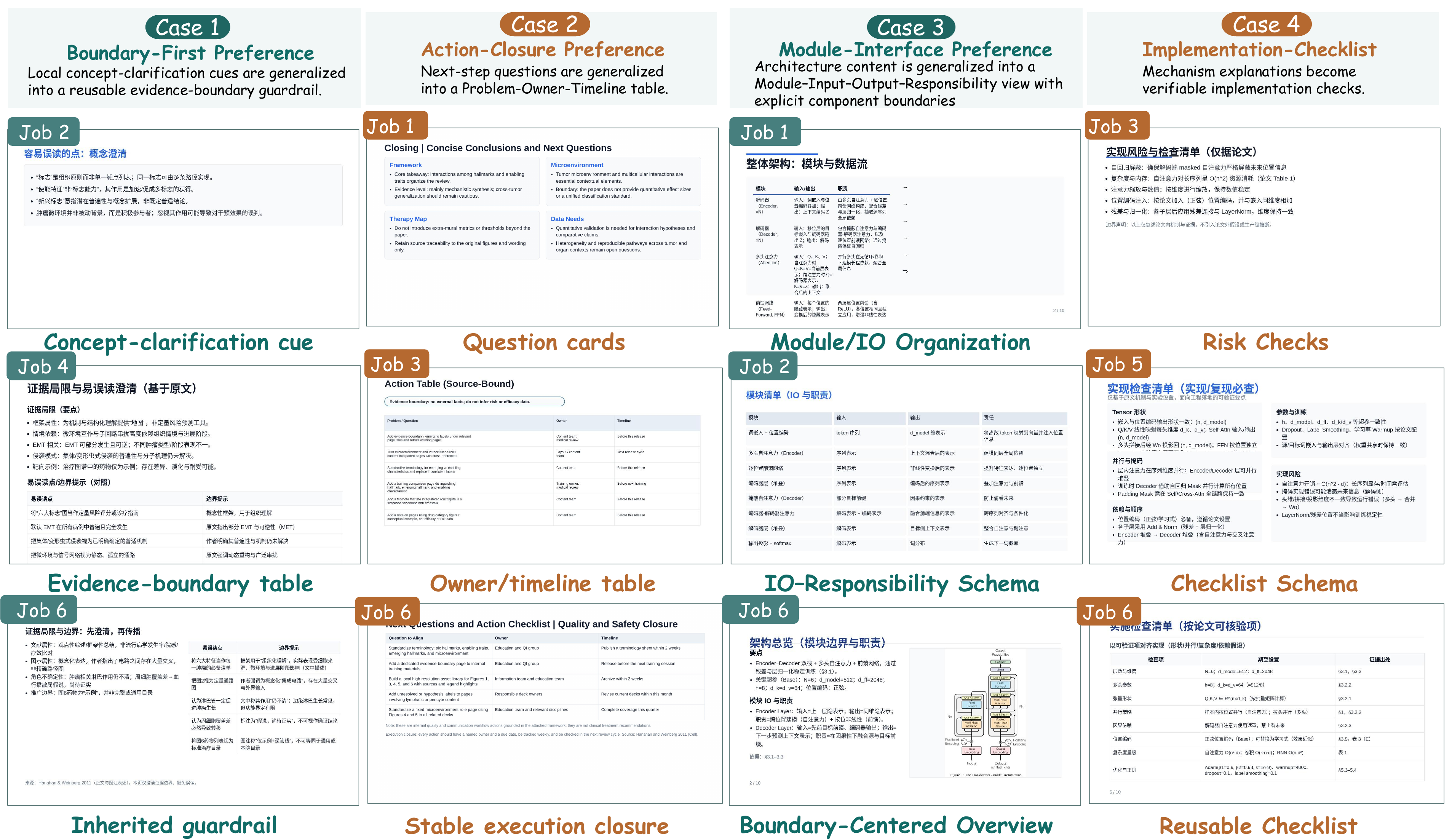}
  \vspace{-2mm}
  \caption{Qualitative cross-job profile consolidation. Across six repeated
  jobs, local feedback cues become reusable profile preferences and later
  reappear as default slide-organization patterns, including evidence-boundary
  guardrails, owner/timeline closure tables, module input--output responsibility
  schemas, and implementation checklists. The figure is diagnostic case evidence
  rather than a separate quantitative metric.}
  \label{fig:six_round_profile_consolidation}
\end{figure}

\noindent\textbf{Localized revision should be judged by locality as well as success.}
Figure~\ref{fig:localized_modify_example} shows the qualitative difference
behind the modify setting: a system may satisfy the requested change while
still touching non-target regions, whereas \methodname keeps the patch closer
to the requested element. This is why the localized-revision metrics are
paired with process constraints such as closed-loop completion, verification,
and core-tool-time efficiency.

\noindent\textbf{Scope of profile and working-memory evidence.}
The round-0 persona-alignment table is the main quantitative evidence for user
profile memory. Figure~\ref{fig:six_round_profile_consolidation} provides
qualitative cross-job profile-consolidation evidence, showing how repeated local
cues become reusable organization preferences in later jobs. Appendix
Figure~\ref{fig:appendix_working_memory_carryover} provides complementary
within-session evidence for delayed carryover of active temporary memory, and
Appendix Figure~\ref{fig:appendix_template_generation_showcase} shows
template-guided generation examples.

\noindent\textbf{Tool memory improves overall reliability and search efficiency, but not by winning every pair.}
In the diagnostic matched-pair modify setting, tool-memory injection raises
overall Closed-Loop Completion from 0.815 to 0.963 (W-L-T-NA 3-1-5-0) and
Strict Verify from 0.310 to 0.534 (8-1-0-0). It also reduces Time to First
Correct Edit from 609.5s to 242.5s (6-2-0-1) and lowers the geometric mean
Core Tool Time Ratio to 0.327$\times$ (8-1-0-0), indicating less
non-inspection tool work. The pair-level counts are important: one Gemini
hard-modify pair loses on Closed-Loop Completion, two pairs lose on first-edit
latency, and one GPT-5 pair uses more core tool time. Together with the
pair-level detail in Appendix~\ref{app:tool_memory_pair_details}, these process
metrics are consistent with a more constrained localized editing pattern rather
than repeated full-deck regeneration or broad exploratory edits.

\FloatBarrier

\section{Conclusion}

We introduced \methodname, a hierarchical memory framework for personalized
presentation generation. By separating user profile memory, active temporary
memory, and tool memory, \methodname supports round-0 persona alignment and
multi-turn localized revision. Controlled experiments show improved persona
alignment and diagnostic gains in local modify reliability.

\section{Limitations}

Our evidence is scoped to controlled persona-alignment judgments, diagnostic
matched-pair modify settings, and qualitative working-memory cases. The profile
bank and edit requests are proxies rather than real-user deployment studies.
Future work should add broader human studies, randomized edit sets, and stronger
memory consent, deletion, and sensitive-preference safeguards.

{
\small
\bibliographystyle{plainnat}
\bibliography{references}
}

\clearpage
\appendix
\section{Appendix}

The appendix provides protocol details, supplemental quantitative tables, and
qualitative examples that clarify how different memory signals appear in
generated presentations. The qualitative figures are intended as diagnostic
illustrations, while the aggregate claims remain tied to the quantitative
tables in the main text and appendix.

\subsection{Evaluation Protocol Details}
\label{app:evaluation_protocol_details}

Section~4 defines the evaluated claims and reports the main metrics. This
appendix subsection records the protocol controls used to keep those metrics
attributable to the corresponding memory condition, rather than to prompt
leakage, arm ordering, or unmatched source materials.

\paragraph{Blind profile-memory judging.}
The profile-memory evaluation uses complete round-0 decks as the judging unit.
Before judging, every candidate deck is rendered into page-aligned images and
assigned an anonymous arm label. The judge receives only the target persona
summary, the aligned deck images, and the dimension rubric; the original user
prompt, parsed intent, system identity, and memory condition are hidden. This
separation is important because the profile claim concerns whether persistent
memory changes the produced deck, not whether a judge can recover instructions
that were explicitly written in the current prompt. Each persona-alignment
dimension receives three blind votes. We aggregate the votes under anonymous
arm labels and map arms back to systems only after aggregation.

\paragraph{Persona-alignment judgment operationalization.}
The persona-alignment judgments report four 0--10 dimensions. \textit{Content}
measures whether content selection, evidence type, emphasis, and wording match
the target persona. \textit{Structure} measures whether page order and
page-type/layout fit reflect persona-conditioned deck organization; it excludes
template matching because this dimension targets deck organization rather than
template retrieval accuracy. \textit{Visual} measures information density,
whitespace, chart/card/diagram style, visual hierarchy, and overall visual tone.
\textit{Specificity} uses distractor personas to test whether the deck remains
identifiable as the target persona rather than a generic professional
presentation.

\paragraph{Matched tool-memory pairs.}
The tool-memory protocol evaluates localized multi-turn modification as paired
agent executions. Within each pair, the source deck, model family, persona, and
modify request are fixed; the compared condition is whether tool memory is
injected into the run. The reported process metrics are computed from recorded
execution traces: edit completion, verification after change, finalization,
time to first correct edit, and core tool time. Inspection and
markdown-conversion calls are excluded from core tool time because they reflect
observation and format transfer rather than edit execution. The modify
scenarios are selected before inspecting pair outcomes and are intended as a
diagnostic controlled setting, not as a distribution-level estimate over all
possible user revisions. 
\paragraph{Tool-metric operationalization.}
Table~\ref{tab:tool_memory_table1_redesign} uses two higher-is-better metrics
and two lower-is-better efficiency metrics. \textit{Closed-Loop Completion}
measures whether a modify run reaches a successful local edit, verifies the
edited result, and finalizes the revised deck. \textit{Strict Verify After
Change} measures whether successful slide-changing edits are followed by local
verification within a short tool-call window, so it rewards edit-and-check
behavior rather than late or unrelated inspection. \textit{Time to First Correct
Edit} is the wall-clock interval from modify-task start to the first
slide-changing edit that satisfies the requested change. \textit{Core Tool Time
Ratio} is the geometric mean of memory/no-injection core-tool time ratios after
excluding inspection and markdown-conversion tools, with the no-injection arm
normalized to 1.0. These definitions make the table about the editing
trajectory itself: whether the agent reaches the requested local change, checks
it promptly, and does so with less non-inspection tool work.

\paragraph{Reproducibility information.}
The paper reports the evaluated systems, model families, judge rubrics,
evaluation configurations, matched-pair definitions, and trace-derived metrics
needed to inspect the reported protocol. The source code is publicly available,
and selected evaluation artifacts will be released when documentation and
licensing checks are finalized.
Table~\ref{tab:appendix_eval_protocol_summary} summarizes the main protocol
controls for the profile-memory and tool-memory evaluations.

\begin{table}[!htbp]
\centering
\small
\setlength{\tabcolsep}{5pt}
\caption{Protocol controls used to isolate memory effects in Section~4.}
\label{tab:appendix_eval_protocol_summary}
\begin{tabular}{lll}
\toprule
Control & Profile-memory evaluation & Tool-memory evaluation \\
\midrule
Evaluation unit & Complete round-0 deck & Paired modify execution \\
Matched factors & Source, model family, task prompt & Source deck, model, persona, request \\
Varied condition & Profile-memory injection & Tool-memory injection \\
Hidden signals & Prompt, intent, system, memory label & Memory-condition label \\
Aggregation & Three blind votes per dimension & Pair-level trace metrics \\
Evidence scope & Main persona-alignment result & Diagnostic selected-pair setting \\
\bottomrule
\end{tabular}
    \end{table}

\subsection{Baseline Conditions and Prompt/Profile Separation}
\label{app:baseline_provenance}

The persona-alignment judgment comparison is designed to isolate persistent profile
memory from prompt-level task conditioning. The memory-injected condition and
its matched control use the same source material, model family, target persona,
task prompt, and generation pipeline. The control condition keeps the authoring
task unchanged but withholds long-term profile memory, session preference
memory, and reusable tool experience from the generation context. The
memory-injected condition receives the same task information plus the routed
profile entry for the target persona-intent pair.

This separation matters because the task prompt already contains the
source-specific request and role intent. Profile memory is therefore evaluated
as an additional persistent signal: structured preferences accumulated across
controlled authoring jobs, not a replacement for the user request. SlideTailor is
included as an external reference/template-conditioned personalization baseline;
it is evaluated from its generated presentations under the same judging
protocol, but it is not treated as an internal memory ablation.

The judge-side protocol further prevents prompt or template leakage from
dominating the score. The persona-alignment judge receives the target persona
summary and rendered deck images, but not the original prompt, parsed task
intent, system identity, or memory condition. The Structure metric deliberately
excludes template matching, so the main profile table evaluates
persona-conditioned organization and layout fit rather than template retrieval
accuracy.

\subsection{Compute Resources and Runtime Accounting}
\label{app:compute_resources}

All reported experiments are inference-time agent evaluations; no model
training or fine-tuning is performed. Runs were orchestrated from a local
Linux workstation using Python 3.11.14. The machine has 48 Intel Xeon Silver
4214R CPU threads and 503 GiB of system memory; although the machine also has
NVIDIA A40 GPUs, the reported experiments are primarily API-bound rather than
GPU-bound. The local machine is used for agent orchestration, document
conversion, slide rendering, image inspection, and tool-mediated HTML/PPT
editing; LLM generation and judging are performed through external model APIs.

For runtime accounting, each run records model-response usage, local tool-call
outcomes, tool active time, and first-to-last-event wall-clock span. Because
different LLM providers expose token accounting fields with slightly different
schemas, we use these records as reproducibility accounting rather than as a
normalized price estimate. Table~\ref{tab:appendix_compute_accounting} reports
the resulting accounting summary. The prompt and completion token columns sum
provider-reported token usage; tool-call columns count successful and failed
local tool invocations; tool active time sums local tool execution duration; and
wall-clock span measures the elapsed time from the first recorded event to the
last recorded event within each run.

\begin{figure}[!t]
  \centering
  \includegraphics[width=0.95\linewidth]{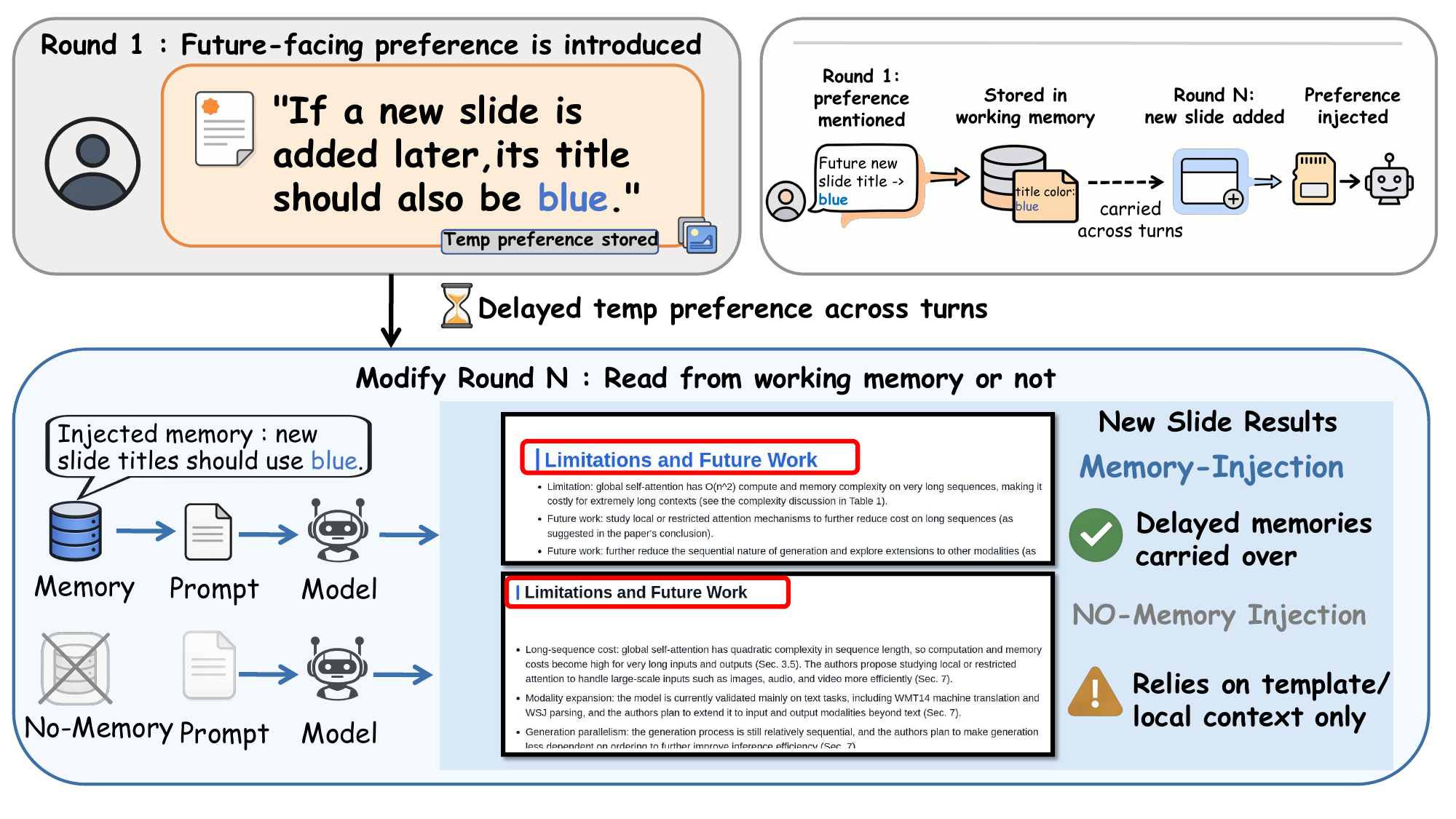}
  \par\vspace{2pt}
  \textbf{(a) Title-color carryover.}
  \par\vspace{8pt}

  \includegraphics[width=0.95\linewidth]{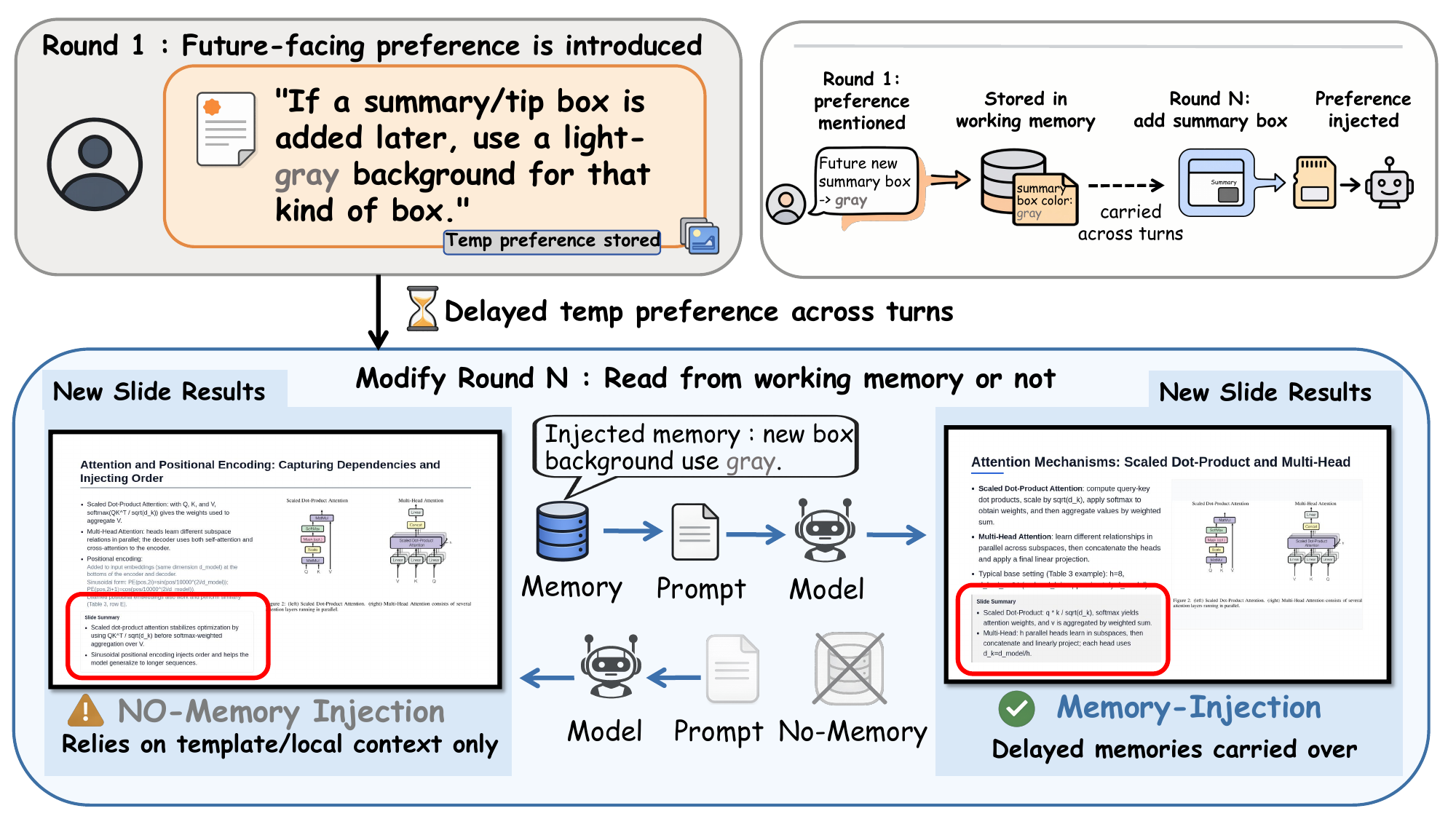}
  \par\vspace{2pt}
  \textbf{(b) Summary-box style carryover.}
  \caption{Working-memory cases for delayed preference carryover. A
  future-facing rule is stated in an earlier turn and only becomes actionable
  after a later edit. Memory injection retrieves the stored rule at the trigger
  turn, whereas the no-memory setting relies mainly on local context.}
  \label{fig:appendix_working_memory_carryover}
\end{figure}

The accounting rows are evidence scopes rather than additional experimental
conditions. The profile-memory row summarizes the generation and analysis runs
used for profile-memory experiments, so it should not be read as the number of
decks used in Table~\ref{tab:profile_memory_v6_bestof_main}.
The tool-memory row corresponds to the diagnostic matched-pair setting reported
in Tables~\ref{tab:tool_memory_table1_redesign} and
\ref{tab:tool_memory_table1_pair_detail}, counting both the memory-injected and
no-injection arms for the nine matched pairs. The working-memory row covers the
four runs used to construct the two qualitative carryover cases in
Figure~\ref{fig:appendix_working_memory_carryover}. The table is intended to
document API budget and local orchestration/runtime footprint, not to serve as a
method-performance result or a GPU-compute claim.

\begin{table}[!htbp]
\centering
\scriptsize
\setlength{\tabcolsep}{3pt}
\renewcommand{\arraystretch}{1.05}
\caption{API usage and local runtime for reported evidence scopes.}
\label{tab:appendix_compute_accounting}
\begin{tabular*}{\textwidth}{@{\extracolsep{\fill}}lrrrrcrrr@{}}
\toprule
Evidence scope & Runs & \shortstack{LLM\\calls} &
\shortstack{Prompt\\tokens} & \shortstack{Completion\\tokens} &
\shortstack{Tool calls\\ok/err} & \shortstack{Tool active\\time (min)} &
\shortstack{Wall span\\sum (h)} & \shortstack{Median run\\span (min)} \\
\midrule
Profile-memory experiments & 49 & 2{,}257 & 49.82M & 2.74M & 3{,}527/552 & 60.78 & 19.37 & 19.12 \\
Tool-memory nine-pair & 18 & 1{,}085 & 19.40M & 0.92M & 1{,}931/220 & 62.25 & 17.52 & 43.99 \\
Working-memory cases & 4 & 161 & 2.35M & 0.27M & 343/21 & 13.85 & 2.64 & 35.09 \\
\bottomrule
\end{tabular*}
\end{table}

For the diagnostic matched-pair tool-memory setting, the recorded process metrics
also record the core tool-time used in Table~\ref{tab:tool_memory_table1_redesign}:
110.5 seconds for memory-injected runs and 354.8 seconds for no-injection runs,
after excluding inspection and markdown-conversion tools. Under the same
exclusion, the matched runs contain 779 core tool calls for memory-injected
runs and 878 for no-injection runs.

\subsection{Profile-Bank Construction}
\label{app:profile_bank_construction}

The profile bank is built as a controlled multi-persona, multi-intent
testbed for long-term personalization. We define ten occupation-style persona
profiles, each associated with three role-intent buckets, resulting in 30
persona-intent profile entries. The personas cover postsecondary teacher,
software developer, management analyst, marketing manager, graphic designer,
training and development specialist, financial manager, operations manager,
medical and health services manager, and legislator. Each entry is used as the
read/write unit for long-term profile memory during generation.
Table~\ref{tab:appendix_profile_library_catalog} lists all 30 entries in the
completed profile bank. The third column reports shortened English summaries
of stored \texttt{profile\_json} fields: \texttt{layout.slide\_structure},
\texttt{visual.chart\_type\_priority}, and, where useful,
\texttt{content.notes}.

\begingroup
\small
\setlength{\tabcolsep}{2.4pt}
\renewcommand{\arraystretch}{1.05}
\setlength{\LTleft}{0pt}
\setlength{\LTright}{0pt}
\setlength{\LTcapwidth}{\textwidth}
\begin{longtable}{>{\scriptsize\raggedright\arraybackslash}p{0.22\linewidth}
                  >{\scriptsize\raggedright\arraybackslash}p{0.25\linewidth}
                  >{\scriptsize\raggedright\arraybackslash}p{0.43\linewidth}}
\caption{Complete 30-entry profile bank used for long-term
personalization. Each row corresponds to one persona-intent profile entry. The
third column reports English summaries of stored \texttt{profile\_json} fields;
long values are shortened for readability.}
\label{tab:appendix_profile_library_catalog}\\
\toprule
Persona & Role-intent bucket & Stored profile fields (English summaries) \\
\midrule
\endfirsthead
\toprule
Persona & Role-intent bucket & Stored profile fields (English summaries) \\
\midrule
\endhead
\bottomrule
\endfoot
Postsecondary teacher & Course unit teaching &
\texttt{layout.slide\_structure}: cover $\rightarrow$ learning objectives
$\rightarrow$ core concept breakdown $\rightarrow$ method workflow
$\rightarrow$ classroom example $\rightarrow$ recap/Q\&A;
\texttt{content.notes}: organize course units around objectives, concepts,
methods, and examples. \\
Postsecondary teacher & Research method explanation &
\texttt{layout.slide\_structure}: two-column layout with explanation on the
left and architecture/local diagrams on the right;
\texttt{content.notes}: for graduate students, clarify problem definition,
method steps, key experimental settings, and method rationale. \\
Postsecondary teacher & Academic progress review &
\texttt{layout.slide\_structure}: background/goals---stage progress---key
results---open issues---next plan;
\texttt{content.notes}: include core parameterized formulas and emphasize
precise academic definitions. \\
\midrule
Software developer & Architecture walkthrough &
\texttt{layout.slide\_structure}: overview/architecture diagram
$\rightarrow$ module structure $\rightarrow$ data flow $\rightarrow$ design
trade-offs $\rightarrow$ experimental evidence;
\texttt{visual.chart\_type\_priority}: architecture/topology diagrams for
overview pages and flowcharts for data flow. \\
Software developer & Sprint risk update &
\texttt{layout.slide\_structure}: current goal $\rightarrow$ completed
capabilities $\rightarrow$ main risks $\rightarrow$ key dependencies
$\rightarrow$ next steps;
\texttt{visual.chart\_type\_priority}: status-risk comparison diagrams and
milestone timelines. \\
Software developer & Technical value brief &
\texttt{content.notes}: organize into application scenarios, core technical
capabilities, performance evidence, and engineering value;
\texttt{visual.chart\_type\_priority}: comparison tables and evidence charts. \\
\midrule
Management analyst & Organizational diagnosis brief &
\texttt{layout.slide\_structure}: current issues---root-cause breakdown---capability
gaps---optimization directions;
\texttt{visual.chart\_type\_priority}: issue trees, $2\times2$ priority
matrices, capability heatmaps, and fishbone diagrams. \\
Management analyst & Framework training session &
\texttt{layout.slide\_structure}: use structured summary pages such as
$2\times2$ matrices or card frameworks with action orientation;
\texttt{visual.chart\_type\_priority}: process diagrams, decision trees, and
priority matrices. \\
Management analyst & Recommendation follow-up review &
\texttt{layout.slide\_structure}: multi-column card layout and priority
matrix for follow-up actions;
\texttt{content.notes}: prioritize conclusions and logic chains over raw data
stacking. \\
\midrule
Marketing manager & Go-to-market pitch &
\texttt{layout.slide\_structure}: open with a hook, use numeric anchors in the
middle, and close with a call to action;
\texttt{content.notes}: prefer short, forceful titles and remove redundant
labels such as ``objective.'' \\
Marketing manager & Brand campaign showcase &
\texttt{layout.slide\_structure}: put the core narrative on the cover; use big
numbers and conclusion cards on inner slides while reducing tables;
\texttt{content.notes}: avoid meta-labels on the cover. \\
Marketing manager & Campaign performance review &
\texttt{layout.slide\_structure}: highlight key metrics with KPI cards and
reduce table/screenshot clutter;
\texttt{visual.chart\_type\_priority}: KPI cards and metric cards, avoiding
table screenshots. \\
\midrule
Graphic designer & Visual direction showcase &
\texttt{layout.slide\_structure}: cover with theme/signature color
$\rightarrow$ mood board $\rightarrow$ color palette and usage ratio
$\rightarrow$ font pairing and type scale $\rightarrow$ grid/spacing rules;
\texttt{content.notes}: avoid information overload and keep the visual
direction sparse and clear. \\
Graphic designer & Design rationale pitch &
\texttt{layout.slide\_structure}: cover $\rightarrow$ design goals/audience
$\rightarrow$ constraints and brand tone $\rightarrow$ alternatives/trade-offs
$\rightarrow$ visual rationale;
\texttt{visual.chart\_type\_priority}: annotated design diagrams, trade-off
matrices, scorecards, and before/after cards. \\
Graphic designer & Design iteration review &
\texttt{layout.slide\_structure}: emphasize clear typography hierarchy and
avoid overlap between text, graphics, and containers;
\texttt{visual.chart\_type\_priority}: side-by-side before/after cards and
annotated mockups with numbered callouts. \\
\midrule
Training and development specialist & Capability training workshop &
\texttt{layout.slide\_structure}: organize by numbered learning-path steps;
\texttt{content.notes}: use action-oriented titles with step numbers and cover
objectives, key steps, examples, and recap reminders. \\
Training and development specialist & Learning solution pitch &
\texttt{layout.slide\_structure}: consolidate information into three or four
logical modules and reduce scattered bullets;
\texttt{content.notes}: prefer diagrammatic expression while keeping key terms
and conclusion prompts such as expert recommendations. \\
Training and development specialist & Rollout effectiveness review &
\texttt{layout.slide\_structure}: organize around training goals, execution
coverage, effectiveness observations, key issues, and improvement plans;
\texttt{visual.chart\_type\_priority}: process diagrams, highlighted tables,
and comparison tables. \\
\midrule
Financial manager & Budget and ROI assessment &
\texttt{layout.slide\_structure}: one table per slide for comparison/impact
tables, avoiding table walls; organize around inputs, expected returns, key
metrics, and constraints;
\texttt{visual.chart\_type\_priority}: key-variable impact tables and KPI
summary tables. \\
Financial manager & Finance control training &
\texttt{layout.slide\_structure}: organize around policy goals, key controls,
common risks, and execution requirements;
\texttt{visual.chart\_type\_priority}: comparison tables, checklists, process
diagrams, and KPI scorecards. \\
Financial manager & Financial risk review &
\texttt{layout.slide\_structure}: use conclusion/risk-stance titles and risk
indicator panels with changes in liquidity, leverage, and coverage;
\texttt{content.notes}: prefer quantitative indicators over vague wording. \\
\midrule
Operations manager & Execution alignment plan &
\texttt{layout.slide\_structure}: prioritize process flow, combine timelines
and kanban boards, and keep work-package boundaries clear;
\texttt{visual.chart\_type\_priority}: flowcharts, timelines, and kanban
boards. \\
Operations manager & Process adoption training &
\texttt{layout.slide\_structure}: convert lists into structured visuals:
horizontal process diagrams for steps, responsibility matrices for ownership,
and card grids for risks;
\texttt{visual.chart\_type\_priority}: horizontal flowcharts, card grids, and
responsibility matrices. \\
Operations manager & Delivery risk review &
\texttt{layout.slide\_structure}: use kanban/card layouts for baseline and
configuration points; use tables for risk-remedy mapping;
\texttt{content.notes}: cover delivery goal, current state, blocking risks,
impact scope, and remediation plan. \\
\midrule
Medical and health services manager & Clinical workflow evidence review &
\texttt{layout.slide\_structure}: structured ten-slide flow around workflow
status, key evidence, metric changes, and improvement directions;
\texttt{visual.chart\_type\_priority}: metric dashboards, horizontal clinical
pathway diagrams, alert cards, and risk panels. \\
Medical and health services manager & Compliance protocol training &
\texttt{layout.slide\_structure}: use tables for checklists and comparisons,
especially two-column tables in right-side modules;
\texttt{content.notes}: clearly distinguish investigational from approved
status for drugs or treatments. \\
Medical and health services manager & Quality and safety review &
\texttt{layout.slide\_structure}: visual-first layout with large metric cards,
comparison diagrams for risks, and flowcharts for processes;
\texttt{content.notes}: organize quality/safety reviews around key metrics,
incidents, cause analysis, corrective actions, and next steps. \\
\midrule
Legislator & Public issue evidence brief &
\texttt{layout.slide\_structure}: ten-slide evidence brief organized as issue
background---core evidence---policy options---public impact;
\texttt{visual.chart\_type\_priority}: comparison tables for policy trade-offs
and tabular evidence. \\
Legislator & Policy communication briefing &
\texttt{layout.slide\_structure}: policy goals $\rightarrow$ key content
$\rightarrow$ implementation points $\rightarrow$ public understanding path;
\texttt{visual.chart\_type\_priority}: evidence-linkage tables. \\
Legislator & Implementation accountability review &
\texttt{layout.slide\_structure}: cover---commitments/framing---implementation
progress and deviations---issue list and impact---accountability---remediation
and supervision---milestone timeline---decision points;
\texttt{visual.chart\_type\_priority}: responsibility matrices, issue-owner
tracking tables, and compact remediation Gantt charts. \\
\end{longtable}
\endgroup

Library construction has two stages. First, each persona-intent entry receives
controlled authoring interactions over the same source material but with
different role-intent prompts, producing initial profile evidence for that
entry. Second, because some accumulated entries remain sparse in structured
fields, we apply a seeded completion step. This step uses stable persona
prompts, an occupation-grounded role-preference registry, and existing profile
signals to fill missing theme, visual, layout, content, and general fields. It
follows only-fill-empty and intent-preservation rules: existing intent-specific
signals are preserved when they already contain usable information, and generic
task constraints such as page count or attachment-use instructions are filtered
out.

The seeded completion is treated as profile-bank construction, not as
additional interaction evidence. It updates only the structured profile entry
and does not create synthetic interaction episodes, tool experiences, or
template-usage records. Each filled field is accompanied by provenance tags
that identify whether it came from the seed prompt, role-preference registry,
current profile signal, or suite intent definition. At runtime, the generation
condition reads the completed profile entry matching the current persona and
role intent, while the no-injection baseline uses the same source task without
profile memory. The evaluation is defined by semantic persona and role-intent
conditions rather than by implementation-specific profile identifiers.

\subsection{DeepPresenter-Style Quality Evaluation}
\label{app:deeppresenter_quality_eval}

We evaluate the generated decks with the general-quality metric family reported
by DeepPresenter~\citep{zheng2026deeppresenter}. This protocol is not a
persona-specific alignment judge and does not replace
Table~\ref{tab:profile_memory_v6_bestof_main} in the main text. Instead, it
asks whether user-profile-memory injection preserves or improves standard
presentation-quality dimensions while improving persona alignment.

The metrics are reproduced as follows. \textit{Constraint}, \textit{Content},
and \textit{Style} are reported on a 1--5 scale. \textit{Content} and
\textit{Style} are computed with the released PPTEval-style prompt
implementation used for presentation quality evaluation~\citep{zheng2025pptagent}.
Content evaluates slide-level content clarity and image-text complementarity,
while Style evaluates slide-level visual appeal and style coherence.
\textit{Constraint} is reimplemented from the DeepPresenter paper definition
for this task suite: it checks hard task constraints and instruction-level
requirements that can be verified from the generated deck and source material.
For SlideTailor, we report a language-adjusted Constraint score because its
current generation setup does not target the requested output language; the
language item is excluded while the remaining hard and semantic checks are
retained. \textit{Diversity} follows the DeepPresenter diversity definition,
using deck-level visual embeddings from DINOv2~\citep{oquab2023dinov2} and the
normalized Vendi score~\citep{friedman2023vendi}; it is a suite-level metric.
\textit{Avg.} is the arithmetic mean of Constraint, Content, and Style, and
Diversity is not included in Avg. All values are re-evaluated on the generated
presentations used in this work.

Table~\ref{tab:profile_memory_v6_ppteval_main} in the main text reports these
general-quality results; bold marks the best score in each metric column over
the full table. Across the three model families, user-profile-memory injection
keeps \textsc{Ours} competitive on the general-quality metrics while the
persona-alignment table shows stronger personalization gains.
The most stable pattern is on Content: \textsc{Ours} is above SlideTailor for
all three model families and is close to or above the DeepPresenter no-injection
baseline. \textsc{Ours} also achieves the highest Avg. on GPT-5 and GLM-5, while
Gemini 3.1 Pro has a lower Avg. because of the Constraint score. On the visual
side, \textsc{Ours} obtains the best Style and Diversity scores for Gemini 3.1
Pro and remains close to the strongest systems in the other model families.
Taken together, the table suggests that user-profile-memory injection is
compatible with competitive general-quality behavior, but not uniformly dominant
on every metric or every model.

This quality check matters because it shows that the main persona-alignment
gains do not come from a trivial trade-off that damages ordinary presentation
quality. The evaluation uses the DeepPresenter quality family as an external
consistency check, with Content and Style from the released PPTEval-style
prompts, Constraint reimplemented from the DeepPresenter-style definition for
this task suite, and Diversity computed from deck-level DINOv2 embeddings with a
normalized Vendi score. It therefore supports a narrow and reviewer-safe
conclusion: MemSlides improves persona-focused generation while remaining
competitive on standard quality metrics, even though the gains are not monotonic
across every model-metric pair.

\subsection{Additional Persona-Alignment Judgment Results}

The GPT-5 profile-memory details indicate that the main persona-alignment trend is
not driven by a single persona. Table~\ref{tab:profile_memory_v6_bestof_gpt_appendix}
reports ten-persona results under the persona-alignment judgments. On
average, profile-memory injection improves Overall alignment by 2.42 points,
with gains of 3.30 on Content, 2.30 on Structure, 3.17 on Visual, and 2.43 on
Specificity. The gains are strongest when the target persona changes evidence
selection, narrative organization, or visual treatment substantially; they are
smaller when the no-injection baseline already happens to match the target
persona reasonably well.

\begin{table}[!htbp]
\centering
\scriptsize
\setlength{\tabcolsep}{3pt}
\caption{GPT profile-memory details under the persona-alignment judgments. Each
metric reports {Ours}, DeepPresenter, and their difference
($\Delta$ = {Ours} - DeepPresenter) on the 0--10 judge scale.
\textit{Overall} averages \textit{Content}, \textit{Structure}, and
\textit{Visual}; \textit{Specificity} is reported separately.}
\label{tab:profile_memory_v6_bestof_gpt_appendix}
\resizebox{\textwidth}{!}{
\begin{tabular}{lccccccccccccccc}
\toprule
Persona & \multicolumn{3}{c}{Overall} & \multicolumn{3}{c}{Content} & \multicolumn{3}{c}{Structure} & \multicolumn{3}{c}{Visual} & \multicolumn{3}{c}{Specificity} \\
\cmidrule(lr){2-4}\cmidrule(lr){5-7}\cmidrule(lr){8-10}\cmidrule(lr){11-13}\cmidrule(lr){14-16}
& {Ours} & DeepPresenter & $\Delta$ & {Ours} & DeepPresenter & $\Delta$ & {Ours} & DeepPresenter & $\Delta$ & {Ours} & DeepPresenter & $\Delta$ & {Ours} & DeepPresenter & $\Delta$ \\
\midrule
\textbf{Average (10)} & \textbf{7.70} & \textbf{5.28} & \textbf{+2.42} & \textbf{8.35} & \textbf{5.05} & \textbf{+3.30} & \textbf{8.07} & \textbf{5.77} & \textbf{+2.30} & \textbf{7.97} & \textbf{4.80} & \textbf{+3.17} & \textbf{7.13} & \textbf{4.70} & \textbf{+2.43} \\
\midrule
Graphic designer & 7.33 & 3.33 & +4.00 & 9.00 & 2.33 & +6.67 & 7.33 & 3.33 & +4.00 & 8.33 & 3.33 & +5.00 & 6.00 & 2.00 & +4.00 \\
Marketing manager & 5.95 & 2.78 & +3.17 & 8.33 & 3.67 & +4.67 & 9.00 & 4.00 & +5.00 & 8.17 & 4.33 & +3.83 & 8.00 & 5.33 & +2.67 \\
Financial manager & 8.33 & 4.33 & +4.00 & 8.33 & 4.67 & +3.67 & 8.67 & 5.33 & +3.33 & 8.50 & 4.00 & +4.50 & 8.00 & 5.00 & +3.00 \\
Postsecondary teacher & 8.67 & 5.33 & +3.33 & 9.00 & 6.00 & +3.00 & 9.00 & 6.67 & +2.33 & 9.00 & 3.33 & +5.67 & 8.67 & 5.33 & +3.33 \\
Operations manager & 7.33 & 6.33 & +1.00 & 8.00 & 3.33 & +4.67 & 8.67 & 6.00 & +2.67 & 8.00 & 3.67 & +4.33 & 6.00 & 4.33 & +1.67 \\
Management analyst & 8.33 & 5.33 & +3.00 & 8.67 & 5.67 & +3.00 & 7.00 & 5.67 & +1.33 & 9.00 & 5.00 & +4.00 & 8.00 & 5.67 & +2.33 \\
Training and development specialist & 9.00 & 6.00 & +3.00 & 9.00 & 5.33 & +3.67 & 9.00 & 5.67 & +3.33 & 8.33 & 4.67 & +3.67 & 8.67 & 5.33 & +3.33 \\
Legislator & 7.40 & 6.67 & +0.73 & 6.67 & 7.00 & -0.33 & 8.33 & 6.33 & +2.00 & 7.67 & 6.00 & +1.67 & 6.67 & 4.00 & +2.67 \\
Software developer & 8.33 & 6.00 & +2.33 & 9.00 & 6.33 & +2.67 & 6.67 & 8.33 & -1.67 & 6.33 & 7.00 & -0.67 & 7.33 & 7.33 & +0.00 \\
Medical health services manager & 6.33 & 6.67 & -0.33 & 7.53 & 6.17 & +1.37 & 7.00 & 6.33 & +0.67 & 6.33 & 6.67 & -0.33 & 4.00 & 2.67 & +1.33 \\
\bottomrule
\end{tabular}
}
\end{table}

\FloatBarrier

\subsection{Template-Guided Generation Examples}

\begin{figure}[!t]
  \centering
  \includegraphics[width=1\linewidth]{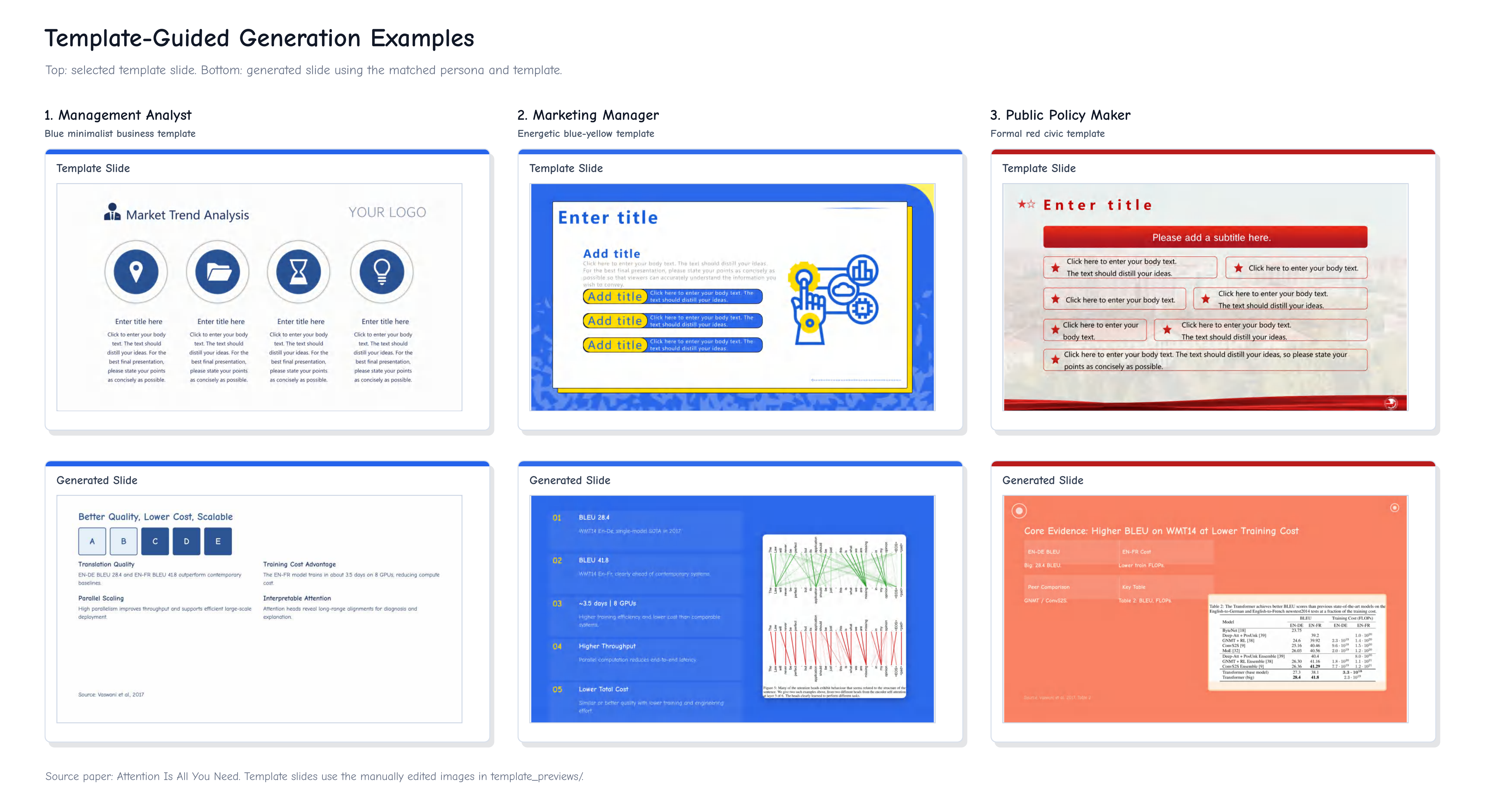}
  \caption{Template-guided generation examples: selected template slides (top)
  and matched persona/template generations (bottom).}
  \label{fig:appendix_template_generation_showcase}
\end{figure}

Figure~\ref{fig:appendix_template_generation_showcase} provides qualitative
examples of template-guided generation under different user roles. Each example
pairs a selected template slide with a generated slide for the same source
paper. The examples show templates acting as concrete task-time design
constraints over layout, palette, typography, and visual organization, while
persona-conditioned generation adapts the slide content and emphasis.

\FloatBarrier

\subsection{Tool-Memory Pair-Level Details}
\label{app:tool_memory_pair_details}

The pair-level tool-memory results make the diagnostic matched-pair setting
transparent. Table~\ref{tab:tool_memory_table1_pair_detail} reports the
tool-memory and no-injection values for each matched pair, together with the
pair verdict from the tool-memory perspective. The Gemini 3.1 Pro rows use a
fixed graphic-designer hard-modify family, so the setting exposes hard cases
rather than selecting pairs by favorable outcomes. The improvements are not from a
single model: most pairs either improve or tie on closed-loop completion, and
most improve Strict Verify. First Correct Edit time and Core Tool Time Ratio
also usually decrease, although not monotonically in every pair. This supports
the diagnostic conclusion that tool memory is associated with more reliable
localized editing behavior, while retaining the caveat that the table summarizes
a diagnostic matched-pair setting.

\begin{table}[!htbp]
\centering
\scriptsize
\setlength{\tabcolsep}{3pt}
\caption{Pair-level details for the diagnostic matched-pair tool-memory setting.
Metric cells follow the same four metrics as Table~\ref{tab:tool_memory_table1_redesign}
and report tool memory / no injection, with verdicts from the tool-memory perspective.}
\label{tab:tool_memory_table1_pair_detail}
\resizebox{\textwidth}{!}{%
\begin{tabular}{lllcccc}
\toprule
Model & Pair & Persona & \shortstack{Closed-Loop\\Completion $\uparrow$} & \shortstack{Strict\\Verify $\uparrow$} & \shortstack{First Correct\\Edit (s) $\downarrow$} & \shortstack{Core Tool\\Time Ratio $\downarrow$} \\
\midrule
GPT-5 & P1 & Graphic Designer & 1.000 / 1.000 (tie) & 0.947 / 0.278 (win) & 264.7 / 317.0 (win) & 0.043$\times$ / 1.000$\times$ (win) \\
GPT-5 & P2 & Management Analyst & 1.000 / 1.000 (tie) & 0.579 / 0.471 (win) & 158.0 / 151.3 (loss) & 18.777$\times$ / 1.000$\times$ (loss) \\
GPT-5 & P3 & Postsecondary Teacher & 1.000 / 0.000 (win) & 0.412 / 0.133 (win) & 93.3 / NA (NA) & 0.502$\times$ / 1.000$\times$ (win) \\
GLM-5 & P4 & Graphic Designer & 1.000 / 0.667 (win) & 0.512 / 0.278 (win) & 128.3 / 215.0 (win) & 0.763$\times$ / 1.000$\times$ (win) \\
GLM-5 & P5 & Graphic Designer & 1.000 / 1.000 (tie) & 0.548 / 0.457 (win) & 177.0 / 224.3 (win) & 0.064$\times$ / 1.000$\times$ (win) \\
GLM-5 & P6 & Graphic Designer & 1.000 / 1.000 (tie) & 0.405 / 0.567 (loss) & 282.3 / 1063.3 (win) & 0.838$\times$ / 1.000$\times$ (win) \\
Gemini 3.1 Pro & P7 & Graphic Designer & 0.667 / 1.000 (loss) & 0.481 / 0.183 (win) & 340.5 / 199.7 (loss) & 0.146$\times$ / 1.000$\times$ (win) \\
Gemini 3.1 Pro & P8 & Graphic Designer & 1.000 / 1.000 (tie) & 0.366 / 0.242 (win) & 329.0 / 2303.0 (win) & 0.254$\times$ / 1.000$\times$ (win) \\
Gemini 3.1 Pro & P9 & Graphic Designer & 1.000 / 0.667 (win) & 0.559 / 0.179 (win) & 260.3 / 402.0 (win) & 0.069$\times$ / 1.000$\times$ (win) \\
\bottomrule
\end{tabular}%
}
\vspace{2pt}
\raggedright
\footnotesize
Aggregating these rows yields Table~\ref{tab:tool_memory_table1_redesign}.
Closed-Loop Completion and Strict Verify use arithmetic means over the three
pairs per model family. First Correct Edit excludes pairs with unavailable
no-injection edit time. Core Tool Time Ratio uses the geometric mean of the
tool-memory/no-injection ratios.
\end{table}

\begin{figure}[!t]
  \centering
  \includegraphics[width=0.9\linewidth]{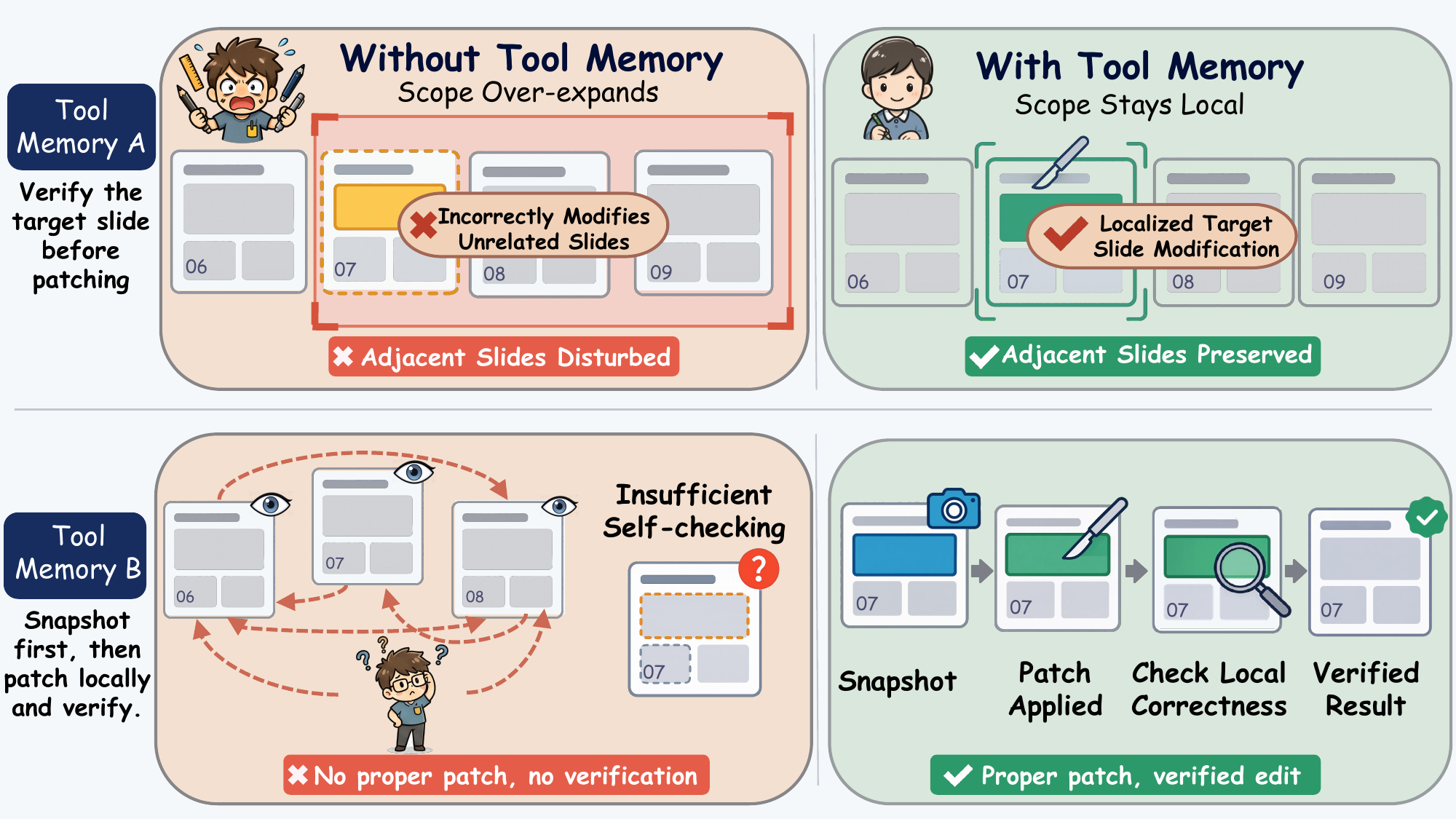}
  \caption{Illustrative qualitative trajectory for localized modify behavior.
  The example contrasts broad or incomplete editing behavior with a more
  constrained trajectory that inspects the target slide, applies a local patch,
  verifies the local change, and finalizes the revision. This figure is a
  process illustration for the diagnostic tool-memory setting rather than a
  separate quantitative result.}
  \label{fig:appendix_tool_memory_localized_editing}
\end{figure}

Figure~\ref{fig:appendix_tool_memory_localized_editing} illustrates the
localized edit trajectory behind these process metrics.

\paragraph{Paired robustness check.}
Table~\ref{tab:tool_memory_paired_robustness} provides a non-parametric
robustness check over the same nine matched pairs. For each metric, we compute
pair-level wins and losses from Table~\ref{tab:tool_memory_table1_pair_detail}
and apply an exact one-sided sign test after excluding ties and unavailable
pairs. The strongest paired evidence appears on Strict Verify and Core Tool
Time Ratio, while Closed-Loop Completion and First Correct Edit remain
directionally favorable. Because the underlying setting is a diagnostic
matched-pair protocol, these checks are intended to characterize robustness
within this controlled protocol rather than to claim significance over the full
distribution of possible modify requests.

The W-L-T-NA column should be read as win--loss--tie--not-available counts from
the tool-memory perspective. A win means that the memory-injected run is better
than its no-injection counterpart under the metric direction: higher is better
for Closed-Loop Completion and Strict Verify, while lower is better for First
Correct Edit time and Core Tool Time Ratio. A tie means the two arms have the
same value. NA means that one arm lacks a comparable value for that metric; for
example, First Correct Edit is unavailable when a run never produces a
verifiable first correct slide edit, so no latency comparison can be made. The
sign test therefore uses only win/loss pairs and asks whether the number of
wins is larger than would be expected under a 50--50 no-effect baseline.

\begin{table}[H]
\centering
\small
\setlength{\tabcolsep}{6pt}
\renewcommand{\arraystretch}{1.03}
\caption{Paired robustness check for the diagnostic matched-pair tool-memory
setting. Win/loss/tie/NA counts are computed from
Table~\ref{tab:tool_memory_table1_pair_detail}. The exact one-sided sign test
uses only wins and losses, excluding ties and unavailable pairs.}
\label{tab:tool_memory_paired_robustness}
\begin{tabular}{lcccc}
\toprule
Metric & W-L-T-NA & Valid W/L & Sign-test $p$ & Interpretation \\
\midrule
Closed-Loop Completion $\uparrow$ & 3-1-5-0 & 4 & 0.3125 & Directional \\
Strict Verify $\uparrow$ & 8-1-0-0 & 9 & 0.0195 & Paired evidence \\
First Correct Edit (s) $\downarrow$ & 6-2-0-1 & 8 & 0.1445 & Directional \\
Core Tool Time Ratio $\downarrow$ & 8-1-0-0 & 9 & 0.0195 & Paired evidence \\
\bottomrule
\end{tabular}
\end{table}

\clearpage

\subsection{Additional Qualitative Memory Cases}

\subsection{Existing Assets and Use Conditions}

We use existing presentation-generation systems, hosted model APIs, visual
representation models, and author-created templates for research comparison,
generation, evaluation, and analysis. We do not redistribute third-party model
weights, proprietary endpoints, commercial services, or third-party code as
part of this submission. Table~\ref{tab:existing_assets} summarizes the main
existing assets used in the experiments and their use conditions.

\begin{table}[!htbp]
\centering
\small
\caption{Existing assets and use conditions.}
\label{tab:existing_assets}
\setlength{\tabcolsep}{5pt}
\begin{tabular}{p{0.24\linewidth}p{0.68\linewidth}}
\toprule
Asset category & Use and condition \\
\midrule
Prior presentation systems &
DeepPresenter, SlideTailor, and PPTAgent are cited and used for baseline
generation, comparison, and evaluation alignment under their released code or
project terms; third-party code is not redistributed. \\
\midrule
Hosted LLMs &
GPT-5, GLM-5, and Gemini 3.1 Pro are used for generation, modification, and
LLM-as-judge evaluation through hosted APIs or institutionally provided endpoints under
applicable provider terms; no model weights or endpoints are redistributed. \\
\midrule
Visual representation models &
DINOv2 is used for deck-level visual embeddings in diversity-style analysis.
The standard DINOv2 model/code release is under Apache License 2.0; checkpoints
are not redistributed. \\
\midrule
Presentation templates &
Author-created templates are used as task-time design constraints for
template-conditioned generation examples and evaluation settings; no
third-party templates are redistributed. \\
\midrule
Evaluation materials &
Profile-memory banks, prompts, generated decks, matched-pair definitions, and
judge outputs are constructed for this paper's controlled evaluation. These
materials are described for reproducibility. The source code is publicly
available, and selected evaluation artifacts will be released when
documentation and licensing checks are finalized. \\
\bottomrule
\end{tabular}
\end{table}

\subsection{Broader Impacts and Responsible Use}

Personalized presentation agents can lower the cost of producing structured
visual communication, helping users translate technical material into decks
that better match their audience, role, and design preferences. Memory
mechanisms may also reduce repeated instruction effort across authoring
sessions and make multi-turn editing more efficient, especially when users
need consistent style, recurring content emphasis, or localized revisions over
an evolving deck.

At the same time, persistent personalization introduces risks. Stored
preference profiles may encode sensitive user habits, organizational style
constraints, or audience-targeting strategies, and incorrect memory
consolidation may preserve outdated or unintended preferences across future
tasks. Presentation generation can also be misused to produce persuasive but
misleading materials, or to over-adapt content framing to a target audience.
These risks suggest that personalized presentation agents should provide
user-visible memory inspection, editing, and deletion controls; avoid storing
unnecessary sensitive information; and separate stable user preferences from
one-off task instructions. Our current work studies controlled generation and
editing settings rather than deployment, and we leave stronger privacy
controls, memory auditing, and real-user governance mechanisms to future work.

\clearpage

\end{document}